%% file: main.tex
\documentclass[conference]{IEEEtran}
\usepackage{times}

\let\labelindent\relax
\usepackage[numbers]{natbib}
\usepackage{multicol}
\usepackage[bookmarks=true]{hyperref}
\usepackage{enumitem}  
\usepackage{float}
\usepackage{afterpage}
\usepackage{stfloats}
\usepackage{tabularx}
\usepackage{array}
\usepackage{makecell}
\usepackage{graphicx}
\usepackage{booktabs}
\usepackage{multirow}
\usepackage{dsfont}
\usepackage{romannum}
\usepackage{hyperref}
\usepackage[dvipsnames]{xcolor}
\usepackage{xspace}
\usepackage{amsmath} % assumes amsmath package installed
\usepackage{amssymb}  % assumes amsmath package installed
\usepackage{cite}
\usepackage{cuted}
\usepackage{caption}
\usepackage{bbm}

\usepackage{mdframed} % for the vertical line
\usepackage[T1]{fontenc} % to ensure good font encoding

\input{text/macros}

\begin{document}

% \input{text/rebuttals} \clearpage

% @inproceedings{ai2024robopack,
  % title={RoboPack: Learning Tactile-Informed Dynamics Models for Dense Packing},
  % author={Bo Ai and Stephen Tian and Haochen Shi and Yixuan Wang and Cheston Tan and Yunzhu Li and Jiajun Wu},
  % booktitle={Proceedings of Robotics: Science and Systems},
  % year={2024}
% }

% \title{RoboPack: Manipulating Objects under Partial Observability with Visuo-Tactile Dynamics Models}
\title{RoboPack: Learning Tactile-Informed \\
Dynamics Models for Dense Packing \vspace{-1.5mm}}

% Potential ideas:
% RoboPack: Learning Visuo-Tactile Dynamics Models for Contact-Rich Manipulation
% 

% You will get a Paper-ID when submitting a pdf file to the conference system
% \author{Author Names Omitted for Anonymous Review. Paper-ID 210}

% \author{\authorblockN{Michael Shell}
% \authorblockA{School of Electrical and\\Computer Engineering\\
% Georgia Institute of Technology\\
% Atlanta, Georgia 30332--0250\\
% Email: mshell@ece.gatech.edu}
% \and
% \authorblockN{Homer Simpson}
% \authorblockA{Twentieth Century Fox\\
% Email: homer@thesimpsons.com}
% \and
% \authorblockN{James Kirk\\ and Montgomery Scott}
% \authorblockA{Starfleet Academy\\
% San Francisco, California 96678-2391\\
% Telephone: (800) 555--1212\\

% avoiding spaces at the end of the author lines is not a problem with
% conference papers because we don't use \thanks or \IEEEmembership

\author{
\authorblockN{\textbf{Bo Ai$^{1,3,4}$\authorrefmark{1}} \quad
\textbf{Stephen Tian$^{1}$\authorrefmark{1}} \quad
\textbf{Haochen Shi$^{1}$} \quad \textbf{Yixuan Wang$^{2}$} \\
\textbf{Cheston Tan$^{3,4}$}  \quad \textbf{Yunzhu Li$^{2}$} \quad
\textbf{Jiajun Wu$^{1}$}}
\authorblockA{\authorrefmark{1}Equal contribution}
\vspace{5pt}

\authorblockA{$^{1}$Stanford University, USA \quad $^{2}$University of Illinois Urbana-Champaign, USA}
% \authorblockA{\authorrefmark{3}Agency for Science, Technology and Research (A*STAR), Singapore}
$^{3}$IHPC, Agency for Science, Technology and Research, Singapore \\
$^{4}$CFAR, Agency for Science, Technology and Research, Singapore
\authorblockA{\textbf{\textcolor{magenta}{\url{https://robo-pack.github.io}}}}
}

% \maketitle
\twocolumn[{%
    \renewcommand\twocolumn[1][]{#1}%
	
        \maketitle
        \vspace{-4mm}
	\begin{center}
		% \includegraphics[width=\textwidth]{figure/UMI-Teaser.pdf}
		% \captionof{figure}
% \makeatletter
% \g@addto@macro\@maketitle{
%   \begin{figure}[H]
%   \setlength{\linewidth}{\textwidth}
%   \setlength{\hsize}{\textwidth}
%   \centering
    \includegraphics[width=\textwidth]{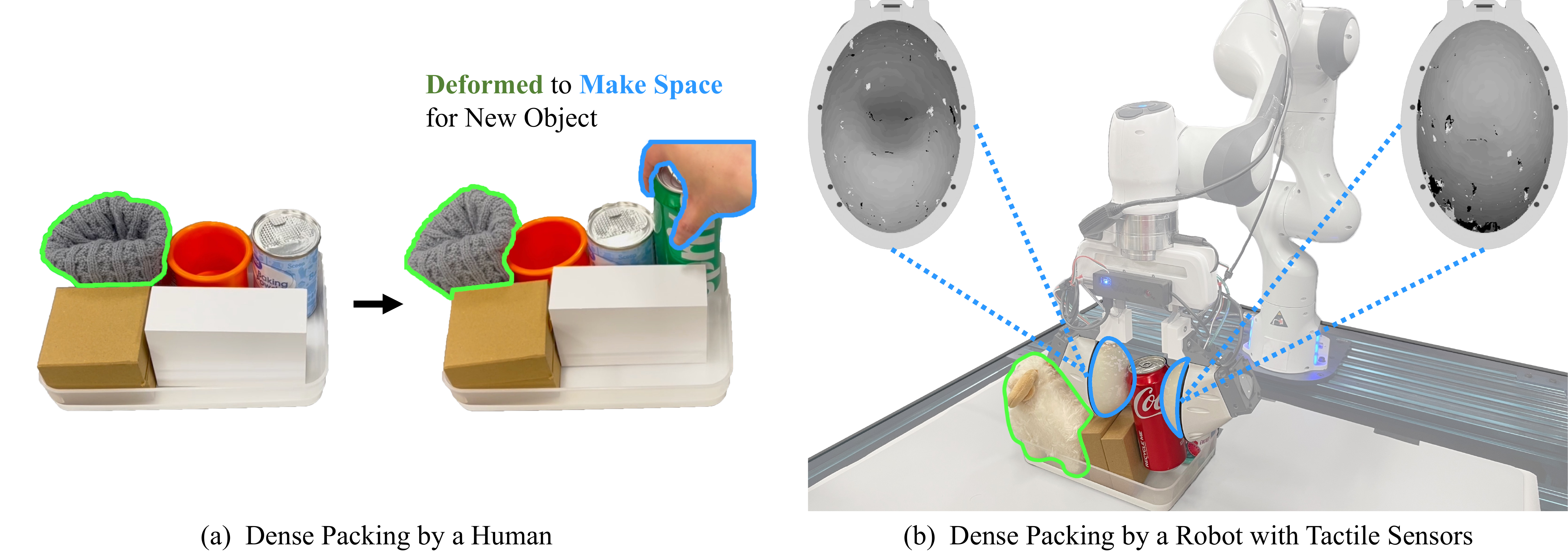}
    \captionof{figure}{
    \textbf{Tactile sensing for dense packing.} Tactile feedback is critical in tasks with heavy occlusion and rich contact, such as dense packing. (a) Humans rely on tactile sensations from their hands to navigate space and fit a water bottle into a suitcase. (b) Likewise, tactile sensing is crucial for robots to perform dense packing tasks, such as placing a can into a packed tray.}
    % \vspace{-0.5cm}
    \label{fig:teaser}
    \end{center}
}]

\input{text/000_abstract.tex}
% \blfootnote{*Equal contribution.}
\input{text/010_intro.tex}

\input{text/020_related-works.tex}
\input{text/030_method.tex}
\input{text/040_setup.tex}
\input{text/050_results.tex}

\input{text/060_conclusion.tex}

%\clearpage
\section*{Acknowledgments}
% We thank the Toyota Research Institute for lending the SoftBubble sensor hardware used in this work. Toyota Research Institute provided funds to support this work. This work is in part supported by the Stanford Human-Centered AI Institute (HAI) and Amazon. ST is supported by NSF GRFP Grant No. DGE-1656518. 

We thank the Toyota Research Institute for lending the SoftBubble sensor hardware. This work is in part supported by the Toyota Research Institute (TRI), the Stanford Human-Centered AI Institute (HAI), and Amazon. S.T. is supported by NSF GRFP Grant No. DGE-1656518. This work is also in part supported by an A*STAR CRF award to C.T.

%
% YL is partially supported by the Toyota Research Institute
%% Use plainnat to work nicely with natbib. 

\bibliographystyle{plainnat}
\bibliography{references}

\clearpage
\input{text/070_appendix.tex}

\end{document}

%% file: text/macros.tex
\newcommand{\secref}[1]{Section~\ref{#1}}
\renewcommand{\eqref}[1]{Eqn~\ref{#1}}
\newcommand{\figref}[1]{Figure~\ref{#1}}

\newcommand{\tabref}[1]{Table~\ref{#1}}

\newcommand{\appendref}[1]{Appendix~\ref{#1}}

\newcommand{\ie}{\textrm{i.e.}}
\newcommand{\eg}{\textrm{e.g.}}

\def\S{\mathcal{S}}
\def\A{\mathcal{A}}

\def\T{\mathcal{T}}
\def\T{\mathcal{J}}
\def\T{\mathcal{T}}
\def\L{\mathcal{L}}
\def\J{\mathcal{J}}
\def\M{\mathcal{M}}
\def\N{\mathcal{N}}
\def\O{\mathcal{O}}

\DeclareMathOperator*{\argmin}{arg\,min}

\newcommand{\rebuttal}[1]{\textcolor{black}{#1}}

\def\method{RoboPack\xspace}
\def\ours{RoboPack\xspace}
\def\oursnotactile{RoboPack (no tactile)\xspace}
\def\gnntactile{RoboCook + tactile\xspace}

\def\simulator{Physics-based simulator\xspace}

\def\MSE{MSE}

\def\pushes{\# Pushes}
\def\SR{Success Rate}

\def\weblink{\urllink[pre = \bgroup\bf, post = \egroup]}

  {\begin{mdframed}[topline=false, bottomline=false, rightline=false,%
    linewidth=1pt,linecolor=black,%
    leftmargin=1cm,rightmargin=1cm,%
    innerleftmargin=10pt,innerrightmargin=10pt,%
    innertopmargin=0pt,innerbottommargin=0pt,%
    skipabove=\baselineskip,skipbelow=\baselineskip]%
   \fontfamily{ppl}\selectfont% Select Palatino font for the quote
  }%
  {\end{mdframed}}

%% file: text/000_abstract.tex
\begin{abstract}

Tactile feedback is critical for understanding the dynamics of both rigid and deformable objects in many manipulation tasks, such as non-prehensile manipulation and dense packing. We introduce an approach that combines visual and tactile sensing for robotic manipulation by learning a neural, tactile-informed dynamics model. Our proposed framework, \method, employs a recurrent graph neural network to estimate object states, including particles and object-level latent physics information, from historical visuo-tactile observations and to perform future state predictions. Our tactile-informed dynamics model, learned from real-world data, can solve downstream robotics tasks with model-predictive control. We demonstrate our approach on a real robot equipped with a compliant Soft-Bubble tactile sensor on non-prehensile manipulation and dense packing tasks, where the robot must infer the physics properties of objects from direct and indirect interactions. Trained on only an average of 30 minutes of real-world interaction data per task, our model can perform online adaptation and make touch-informed predictions. Through extensive evaluations in both long-horizon dynamics prediction and real-world manipulation, our method demonstrates superior effectiveness compared to previous learning-based and physics-based simulation systems. 

\end{abstract}

\IEEEpeerreviewmaketitle

%% file: text/010_intro.tex
\section{Introduction}

Imagine packing an item into a nearly full suitcase. \rebuttal{As humans, we typically} first form a visual representation of the scene and then make attempts to insert the object, \textit{feeling} the compliance of the objects already inside to decide where and how to insert the new object. \rebuttal{If a particular region feels soft, we can then apply additional force to make space and squeeze the new object in. This process is natural for us humans but very challenging for current robotic systems.}

What would it take to produce adept packing capabilities in robots? Firstly, a robot needs to understand how its actions will affect the objects in the scene and how those objects will interact with each other. Dynamics models of the world predict exactly this: how the state of the world will change based on a robot's action. However, most physics-based dynamics models (e.g., physical simulators), assume full-state information and typically exhibit significant sim-to-real gaps, especially in unstructured scenes involving deformable objects.

At the same time, tasks such as dense packing present significant challenges due to severe occlusions among objects, creating partially observable scenarios where vision alone is insufficient to determine the properties of an object, such as its softness, or assess whether there is space for additional objects. For effective operation, the robot must integrate information from its actions and the corresponding tactile sensing into its planning procedure. However, the optimal method for incorporating tactile sensing information into dynamic models is unclear. Na\"ively integrating tactile sensing into a model's state space can perform poorly because the intricate contacts make tactile modeling a challenging problem, as we will also show empirically later on.

To tackle these challenges, in this work, we propose to 1) learn dynamics directly from \textit{real} physical interaction data using powerful deep function approximators, 2) equip our robotic system with a compliant vision-based Soft-Bubble tactile sensor~\citep{kuppuswamy2020softbubble}, and 3) develop a learning-based method for effective estimation of latent physics information from tactile feedback in interaction histories.

Because learning dynamics in raw pixel observation space can be challenging due to the problem's high dimensionality, we instead model scenes using keypoint particles~\citep{manuelli2020keypoints, dpi, shi2023robocooka, shi2022robocraft, shi2023robocraft}. Finding and tracking meaningful keypoint representations of densely packed scenes over time is itself challenging due to the proximity of objects and inter-occlusions. In this work, we extend an optimization-based point tracking system to preprocess raw observation data into keypoints.

We use the Soft-Bubble tactile sensor \citep{kuppuswamy2020softbubble}, which is ideal for tasks like dense packing, as it can safely sustain stress from the handheld object in all directions and provides high-resolution percepts of the contact force via an embedded RGB-D camera.

Finally, we propose an effective way to incorporate tactile information into our system by learning a separate state estimation module that incorporates tactile information from prior interactions and infers latent physics vectors that contain information that may be helpful for future prediction. This allows us to learn \textit{tactile-informed} dynamics.

We call this system comprising keypoint-based perception, latent physics vector and state estimation from tactile information, dynamics prediction, and model-based planning \textbf{\method}. 
We deploy \method on two real-world settings---a tool-use manipulation and a dense packing task. These tasks involve multi-object interactions with complex dynamics that cannot be determined from vision alone. 
Furthermore, these settings are exceptionally challenging because, unlike prior work that only estimates the physical properties of the object held in hand, our tasks also require estimating the physical properties of objects with which the robot interacts \textit{indirectly} through the handheld object. 

We find that our method can successfully leverage histories of visuo-tactile information to improve prediction, with models trained on just 30 minutes of real-world interaction data per task on average. Through empirical evaluation, we demonstrate that \method outperforms previous works on dynamics learning, an ablation without tactile information, and physics simulator-based methods in dynamics prediction and downstream robotic tasks. We further analyze the properties of the learned latent physics vectors and their relationship with interaction history length.

%% file: text/020_related-works.tex
\section{Related Work}

\subsection{Learning Dynamics Models}

Simulators developed to model rigid and non-rigid bodies approximate real-world physics, often creating a significant sim-to-real gap~\citep{weinstein2006dynamic, holl2019learning, murthy2020gradsim}. To address this, we use a graph neural network (GNN)-based dynamics model trained directly on real-world robot interaction data, aligning with data-driven approaches for learning physical dynamics~\citep{nagabandi2020deep, luo2018algorithmic}. Recent works have demonstrated inspiring results in learning the complex dynamics of objects such as clothes~\citep{lin2022learning}, ropes~\citep{chang2020modelbased}, and fluid~\citep{legaard2023constructing}, with various representations including low-dimensional parameterized shapes~\citep{matl2021deformable}, keypoints~\citep{li2020causal}, latent vectors~\citep{kurutach2018learning}, and neural radiance fields~\citep{li20213d}. \method{}, inspired by previous works~\citep{dpi, sanchez-gonzalez2020learning, battaglia2016interaction}, focuses on the structural modeling of objects with minimal assumptions about underlying physics. This approach overcomes the limitations of physics simulators by directly learning from real-world dynamics. Prior work on GNN-based dynamics learning~\citep{shi2022robocraft, shi2023robocooka, shi2023robocraft, wang2023dynamicresolutiond, chen2023predictinga} heavily relies on visual observations for predicting object dynamics, failing to capture unobserved latent variables that affect real-world dynamics, such as object physical properties. To address this challenge, our method incorporates tactile sensing into dynamics learning and leverages history information for state estimation, offering a robust solution to overcome the constraints of vision-only models.
\subsection{\rebuttal{Model-Free and Model-Based Reinforcement Learning}} \label{sec:related-rl}

\rebuttal{{\let\textbf\relax \input{text/rl-related-work}}
}
\subsection{Tactile Sensing for Robotic Manipulation} 
Tactile sensing plays an important role in both human and robot perception~\citep{dahiya2010tactile}. Among all categories of tactile sensors, vision-based sensors such as~\citep{yuan2017gelsighta, donlon2018gelslim, lambeta2020digit, lin20239dtact} can achieve accurate 3D shape perception of their sensing surfaces. In our work, we use the Soft-Bubble tactile sensor~\citep{kuppuswamy2020softbubble} which offers a unique combination of compliance, lightweight design, robustness to continuous contact, and the ability to capture detailed geometric features through high-resolution depth images~\citep{kuppuswamy2020softbubble, suh2022seed}. Previous studies have successfully integrated vision and tactile feedback in robotic manipulation using parallel grippers~\citep{calandra2018more, gao2023object, li2023see} and dexterous hands~\citep{qi2023general, suresh2023neural, yuan2023robot}. In these tasks, vision effectively offers a comprehensive understanding of the scene's semantics, while tactile sensing delivers accurate geometry estimation for objects in contact that are often occluded. In our study, we explore the potential of integrating vision and tactile feedback for learning dynamics in tasks involving rich contact, occlusions, and a diverse set of objects with unknown physical properties, such as box pushing and dense packing.

%% file: text/rl-related-work.tex
Reinforcement learning (RL) aims to derive policies directly from interactions. Our method contrasts with model-free RL approaches \citep{dqn, ddpg, sac, mt-opt, e2e_visuomotor}, by incorporating an explicit dynamics model, enhancing interpretability and including structured priors for improved generalization. \textbf{Our work is closer to model-based RL \citep{texplore, planet, nagabandi2020deep, rambo, bremen, solar}} in that we combine learned world models with planning via trajectory optimization. In particular, we learn world models in an offline manner from pre-collected interaction data, avoiding risky trial-and-error interactions in the real world. \textbf{However, our approach is different from existing offline model-based RL \citep{lompo, daydreamer, visual-foresight, tian2019manipulation, offline_model_rl_survey} as it leverages multiple sensing modalities}, \ie, tactile and visual perception. This multi-modal approach provides a more comprehensive understanding of both global geometry and the intricate local physical interactions between the robot gripper and objects. \textbf{Moreover, our method addresses challenges in scenarios where visual observations are not always available}. It uses tactile observation histories to estimate partially observable states, enabling online adaptation to different dynamics. This integration of offline model learning, multi-modal perception, and online adaptation equips our system with adaptive control behaviors for complex tasks. 

%% file: text/030_method.tex
\begin{figure*}[t]
    \centering
    \includegraphics[width=\textwidth]{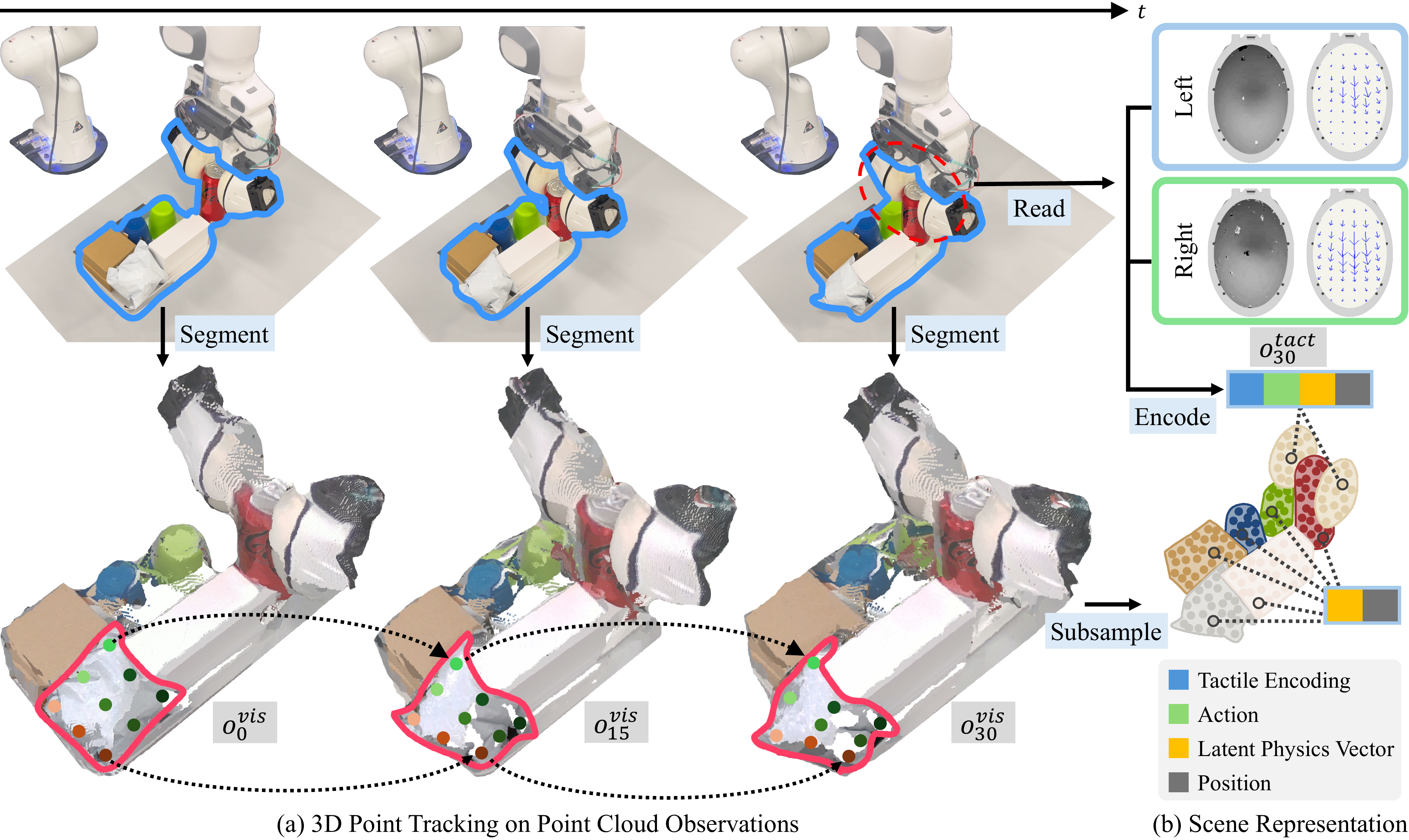}
    \caption{\textbf{\ours's perception module.} (a) We construct a trajectory comprising particle representations of the scene, maintaining correspondence via 3D point tracking on the point cloud data. (b) These particles facilitate the creation of a visual scene representation, denoted as $o^{vis}_t$. For points representing the Soft-Bubble grippers, tactile encodings $o^{tact}_t$ and latent physics vectors are integrated as extra attributes of the particles. We note that while the 3D point tracking module is needed at training time, during deployment the visual feedback can be replaced by predictions from our state estimator. This estimator auto-regressively predicts object particle positions from tactile interaction history and reduces reliance on dense visual feedback, which can be difficult to obtain due to visual occlusions.}
    \vspace{-0.5cm}
    \label{fig:perception}
\end{figure*}

% \section{Problem Statement}
% The objective of this work is to perform robotic manipulation of objects with unknown physics properties under heavy occlusions, which is inherently challenging for robotic systems relying on vision feedback. To formulate the problem, we define the robot's observation space $O$ to include two components: a visual observation $O^{vis}$ and a tactile observation $O^{tact}$. Our goal is to learn the transition function $\T$: $\S \times \A \rightarrow \S$, where the state $s$ is defined as $\langle o^{vis}, \xi \rangle$, where $\xi$ is a latent physics parameter of objects which could include information on mass distribution, friction, and rigidity. The objective is to find a sequence of actions $a_0, a_1, ..., a_{H-1}$ to minimize a cost function $\mathcal{J}$ between the final states and a given target state $s_g$:

% \begin{align}
%     (a_0, ..., a_{H-1}) = \argmin_{a_0, ..., a_{H-1} \in \A{}} \J{}(\T(s_0, (a_0, .., a_{H-1})), s_g).
% \end{align}

% Due to the complexity of modeling dynamics of objects with unknown physical properties, we adopt a data-driven approach to learn the transition function $\T$. With the learned dynamics model, we optimize a future action plan, performing model-predictive control (MPC). 

% The key question we seek to answer is \textbf{how to leverage multi-modal observations to make future predictions on partially observed states}.
\section{Method} \label{sec:method}

\subsection{Overview}
The objective of \method is to manipulate objects with unknown physical properties in environments with heavy occlusions like dense packing. To formulate this problem, we define the observation space as $\O$, the state space as $\S$, and the action space as $\A$. Our goal is to learn a state estimator $g$ that maps $\O$ to $\S$ and a transition function $\T$: $\S \times \A \rightarrow \S$.

% where the state $s$ is defined as $\langle o^{vis}, \xi \rangle$, where $\xi$ is a latent physics parameter of objects which could include information on mass distribution, friction, and rigidity.

To efficiently learn dynamics from real-world multi-object interaction data, we would like to extract lower-dimensional representations of observations like keypoints. Furthermore, we require a mechanism to fuse tactile interaction histories into these representations without full tactile future prediction. Finally, to solve real robotic tasks, we need to leverage our learned model to plan robot actions.

Thus, our system has four main components: perception, state estimation, dynamics prediction, and model-predictive control, discussed in Section~\ref{sec:perception},~\ref{sec:estimator},~\ref{sec:dynamics}, and~\ref{sec:control} respectively. They are used together in the following way: 

First, the perception system extracts particles from the scene as a visual representation $o^{vis}$ and encodes tactile readings into latent embeddings $o^{tact}$ attached to those particles. 

Secondly, the state estimator $g$ infers object states $s$ from any prior interactions, which includes a single visual frame $o^{vis}_0$, the subsequent tactile observations $o^{tact}_{0:t}$, and the corresponding robot actions $a_{1:t-1}$: 
\begin{align}
    \hat{s}_t = g(o^{vis}_0, o^{tact}_{0:t}, a_{1:t-1}).
\end{align}
% where $o^{vis}_0, o^{tact}_{0:t}$ and $a_{1:t-1}$ denote the initial visual observations, tactile observations, and actions respectively. 

Thirdly, to enable model-predictive control, we learn a dynamics prediction model $f$ that predicts future states given the estimated current states and potential actions: 
\begin{align}
    \hat{s}_{t+1} = f(\hat{s_t}, a_t).
\end{align}
Lastly, the future predictions are used to evaluate and optimize the cost of sampled action plans. The objective is to find a sequence of actions $a_0, ..., a_{H-1}$ to minimize a cost function $\mathcal{J}$ between the final states and a given target state $s_g$: 
\begin{align}
    (a_0, ..., a_{H-1}) = \argmin_{a_0, ..., a_{H-1} \in \A{}} \J{}(\T(s_0, (a_0, .., a_{H-1})), s_g).
\end{align}
The robot executes the best actions and receives tactile feedback from the environment, with which it updates its estimates about object properties. 

% Next, we introduce each component in more detail. 

% \begin{enumerate}[label=\roman*.]
%     \item Perception: Our system has a visual and a tactile perception module. For visual perception, we adopt a particle-based object representation and track objects over time with D$^3$Field \citep{wang2023d3fields}, and the point-to-point correspondence from tracking is crucial for providing supervision in model training and to provide intermittent visual feedback during deployment. For tactile perception, we have an auto-encoder that encodes tactile readings from sensors in order to incorporate them into our objects' particle representations. 
%     \item State estimation and dynamics prediction: We propose a recurrent graph neural network \textit{state estimator} to infer the current state based on visual and tactile history observations, and a \textit{dynamics model} that predicts future states given the current estimated state and potential actions. 
%     \item Model-predictive control: We integrate the learned dynamics model with model-predictive control to solve for optimal actions towards a given objective. 
% \end{enumerate}

\subsection{Perception}  \label{sec:perception}
\subsubsection{Visual Perception} 
% \boai{This subsubsection is incomplete and we need a major revision. Would be ideal if we add a couple of equations to explain what it does at the high level, and point out what we added to D3Field, ie, deformation prediction and perhaps mask closeness loss.}
% \stephen{I don't think it's worth the space for the equations, we can put those in the appendix. I don't want the technical details here to get in the way of the main message.}

Our visual perception module extends the formulation of D$^3$Fields~\citep{wang2023d3fields}, with an additional deformation term to handle non-rigid objects and mask-based closeness loss to better support multi-object scenes with occlusion. As shown in Figure~\ref{fig:perception}(a), it takes in multi-view RGB-D observations and outputs tracked 3D keypoints for each object of interest. Critical for our training procedure, these keypoints maintain correspondences over time---a tracked point stays at the same region of an object throughout the trajectory. 

First, we extract visual features for each object with a pre-trained DINOv2 
model \citep{dinov2} and masks using Grounded SAM \citep{dinov2, kirillov2023segany, liu2023grounding}. Through projection and interpolation, we can then compute semantic, instance, and geometric features for arbitrary 3D points. We initialize desired tracking points on object surfaces for an initial frame and formulate 3D keypoint tracking for subsequent frames as an optimization problem. The tracking objective has the following terms:
\begin{itemize}
    \item \textbf{Distance to surface}. Use depth information to encourage points to be close to object surfaces.
    \item \textbf{Semantic alignment}. Align DINOv2 features between projected points in the current and initial frame.
    \item \textbf{Motion regularization}. Penalize large motion between consecutive frames to avoid jitter.
    \item \textbf{Mask consistency}. For multi-object packing settings with significant occlusion, we introduce an objective that constrains tracked points to be near the corresponding object masks, providing more consistent optimization signal for object pose than semantic alignment.
\end{itemize}

We optimize a translation and rotation transformation for each object with this objective. For deformable objects, we also predict axis-aligned shearing scales apart from a rigid transformation to track deformations.

\subsubsection{Tactile Perception} \label{sec:tactile-perception}
As shown in the top right of Figure~\ref{fig:perception}, our tactile perception module takes global force-torque and local force vectors as input and outputs embeddings for the tactile reading. Each Soft-Bubble tactile sensor provides its surface force distribution. This includes (1) shear force vectors $\{ \langle q^x_{i, j}, q^y_{i, j} \rangle \}_{i, j}$, where $i, j$ is the coordinate of a point on the 2D surface of the bubble and $x, y$ denote the vertical and horizontal axis of the tangent plane at that point, as well as (2) a global shear force torque vector and the overall force magnitude $\langle Q^x, Q^y, |Q| \rangle$. $F^x, F^y$ are the mean of local force vectors across spatial dimensions, and $|Q|$ is defined as
% , formally 
% \begin{align}
%     F^x & = \frac{1}{HW} \sum_{i=0}^H \sum_{j=0}^W f^x_{i, j} \\
%     F^y & = \frac{1}{HW} \sum_{i=0}^H \sum_{j=0}^W f^y_{i, j} 
% \end{align}
\begin{align}
    |Q| = \sqrt{ {\max_{i, j} |q^x_{i, j}|}^2 + {\max_{i, j} |q^y_{i, j}|}^2  },
\end{align}
% where $H$ and $W$ are tactile reading's spatial dimensions. 
%The readings are computed and provided by the Soft-bubble gripper API \cite{softbubblegripper}.

\subsubsection{Integrating Visual and Tactile Perception} \label{subsubsec:tactile}

As depicted in Figure~\ref{fig:perception}(b), to integrate tactile observations with particle-based object representation, we first extract particles from the surface of the soft-bubble gripper by projecting the depth camera reading inside the gripper into 3D space. Next, we define a point-wise tactile signal as $\langle q^x_{i, j}, q^y_{i, j}, Q^x, Q^y, |Q| \rangle$ and train an auto-encoder that \rebuttal{maps the point-wise signals independently into latent embeddings. Details regarding the auto-encoder architecture and training are available in \appendref{appendix:ae_details}}. We denote the collection of embeddings as the tactile observation $o^{tact}$. Lastly, we combine the object particles from the visual observation $o^{vis}$ with the tactile sensor particles $o^{tact}$ to form a unified particle representation of the scene. 
 
% In the next subsection, we will describe how to prepare the particles and tactile observations for dynamics learning.

% \subsection{State Estimation and Dynamics Prediction}

\begin{figure}[t]
    \centering \includegraphics[width=\columnwidth]{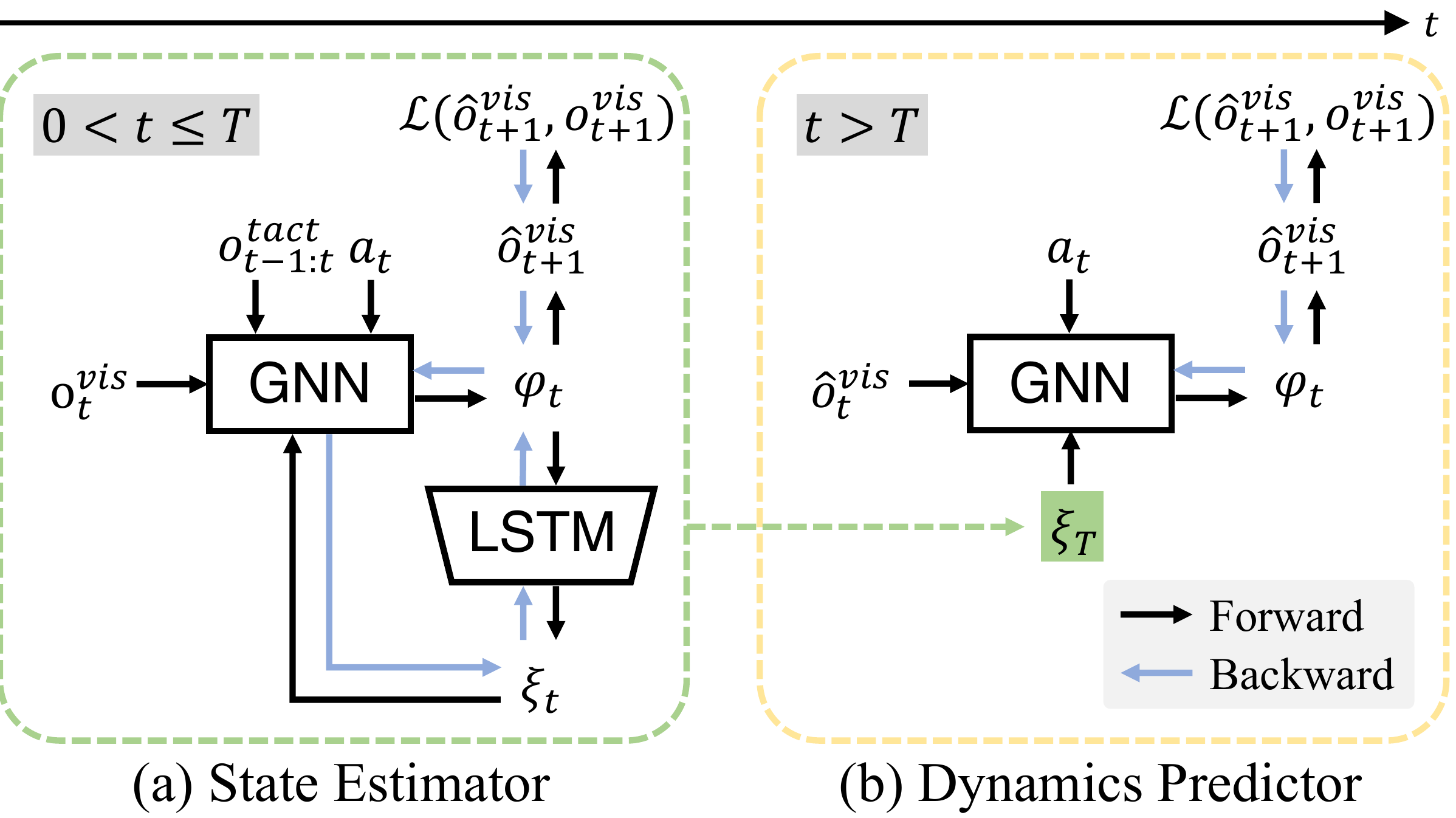}
    \caption{\textbf{\ours's dynamics module.} We perform state estimation and dynamics reasoning with a state estimator and a dynamics predictor respectively. (a) The state estimator auto-regressively predicts the positions of objects' particles and their latent physics vectors, reducing the dependency on dense visual feedback. (b) The dynamics predictor, conditioned on the estimated physics vectors, performs future prediction for planning. These modules share the same architecture, except that the state estimator has an LSTM that integrates history information and predicts physics parameters for each object.}
    \vspace{-0.5cm}
    \label{fig:dynamics}
\end{figure}

\subsection{State Estimation and Latent Physics Vector Inference }\label{sec:estimator}
In real-world robotic manipulation, visual observations are not always available due to occlusion, but knowledge about object dynamics requires interactive feedback. In this work, we leverage tactile feedback to help estimate world states. 

History information is often used to estimate the current state in POMDPs \citep{ai2022deep, pomdp_leslie, sarsop, pomcp}. Similarly, we seek to incorporate tactile history information into state estimation by employing a combination of graph neural networks (GNNs) and long-short term memory (LSTM), as shown in Figure~\ref{fig:dynamics}(a). \rebuttal{We define our state as a tuple of object particles and an object-level latent physics vector, which capture the geometry and physics properties of objects respectively. In the following paragraphs, we describe how our method performs state estimation using history information. }
% recurrent graph neural network for state estimation. 

At time $0<t\leq T$, our state estimator $g$ infers all states for $t = 1, ..., T$ autoregressively. Given the estimated previous state $\hat{s}_{t-1}$ and the tactile feedback at the previous and the current state $o^{tact}_{t-1:t}$, we construct a graph $G_{t-1} = \langle V_{t-1}, E_{t-1} \rangle$ with $V_{t-1}$ as vertices and $E_{t-1}$ as edges. For each node, $v_{i, t-1} = \langle x_{i, t-1}, c^o_{i, t-1} \rangle$, where $x_{i, {t-1}}$ is the particle position $i$ at time $t-1$, and $c^o_{i, t-1}$ are particle attributes. \rebuttal{The particle attributes contain (1) the previous and current tactile readings, $o^{tact}_{t-1:t}$, and (2) the latent physics vector of the object that the particle belongs to, $\xi_{\M{}_i, t-1}$, where $\M_i$ is the object index corresponds to the $i$-th particle, $1 \leq \M_i \leq Z$ and $Z$ is the maximum number of objects in the scene. Formally, $c^o_{i, t-1} = \langle \xi_{t-1}, o^{tact}_{t-1:t} \rangle$. Note that here we implicitly assume that $\M$ is constant (\ie, objects only exhibit elastic and plastic deformations but not break apart), which generally holds for a large number of common manipulation tasks.} Moreover, edges between pairs of particles are denoted $e_k = \langle u_k, v_k \rangle$, where $u_k$ and $v_k$ are the receiver and sender particle indices respectively, and $1 \leq u_k, v_k \leq |V_{t-1}|$ where $k$ is the edge index. We construct graphs by connecting any nodes within a certain radius of each other. 

Given the graph, we first use a node encoder $f^{enc}_V$ and an edge encoder $f^{enc}_E$ to obtain node and edge features, respectively: 
% \begin{align}
    % h^o_{i, t} & = f^{enc}_O(o_{i, t-1}) \label{eqn:object_encode} \\
    % h^e_{k, t} & = f^{enc}_E(e_{i, t-1}) 
% \end{align}
\begin{align}
    h^v_{i, t-1} & = f^{enc}_V(v_{i, t-1}), 
    \label{eqn:object_encode}  & 
    h^e_{k, t-1} & = f^{enc}_E(e_{k, t-1}).
\end{align}

Then, the features are propagated through the edges in multiple steps, during which node effects are processed by neighboring nodes through learned MLPs. We summarize this procedure as $f^{dec}_E$, which outputs an aggregated effect feature for each node called $\phi_i$:

% we use an object decoding function $f^{dec}_V$, a relation function $f^{dec}_E$, and a physics parameter predictor $f^{dec}_\xi$ to estimate the next state and update the physics parameters: 
% \begin{align}
%     b_{k, t} & = f^{dec}_E(h^e_{k, t-1})_{k=1, ..., |E_{t-1}|}  \\
%     \phi_{t-1} & = \sum_{k \in N_i} b_{k, t-1} \\ 
%     \hat{o}_{i, t}^{vis} & = f^{dec}_V\left(h^v_{i, t-1}, \phi_{t-1} \right)_{i=1, ..., |V_{t-1}|} \label{eqn:observation_predict}  \\
%     \xi_{t}, m_{t} & = f^{dec}_\xi\left(h^v_{t-1}, \phi_{t-1}, m_{t-1}\right)  \label{eqn:LSTM}
% \end{align}
% \begin{align}
%      b_{k, t} & = f^{dec}_E(h^e_{k, t-1})_{k=1, ..., |E_{t-1}|}  \\
%      \phi_{i, t-1} & = \sum_{k \in N_i} b_{k, t-1} 
% \end{align}

\begin{align}
     \phi_{i, t-1} & = f^{dec}_E(h^v_{i, t-1}, \sum_{k \in \N_i} h^e_{k, t-1})_{k=1, ..., |E_{t-1}|}.
\end{align}
where $\N_i$ is a set of relations with particle $i$ as the receiver.

Next, the model predicts node (particle) positions and updates the latent physics vector: 
\begin{align}
    \hat{o}_{i, t}^{vis} & = f^{dec}_V\left(h^v_{i, t-1}, \phi_{i, t-1} \right)_{i=1, ..., |V_{t-1}|},\label{eqn:observation_predict}  
\end{align}
\begin{align}
    \xi_{\eta, t}, m_{t} & = f^{dec}_\xi\left(\sum_{\substack{i \\ \M_i = \eta}} h^v_{i, t-1}, \sum_{\substack{i \\ \M_i = \eta}} \phi_{i, t-1}, m_{t-1}\right)_{\eta=1,...,Z}.  \label{eqn:LSTM}
\end{align}
where $f^{dec}_\xi$ is an LSTM, $m_t$ is its internal cell state at the current step, and $\xi_{\eta, t}$ is the updated physics latent vector for $\eta$-th object. At $t = 0$ the LSTM state $m_0$ is initialized as zero. The physics vector for each object is initialized as Gaussian noise: $\xi_{\eta, 0} \sim \mathcal{N}(0, 0.1^2)$ for all $\eta$. \rebuttal{All other encoder and decoder functions (\ie, $f^{enc}_V, f^{dec}_V$, $f^{enc}_E$, and $f^{dec}_E$) are MLPs.}

\subsection{Dynamics Prediction}
\label{sec:dynamics}
% After the state has been estimated, our dynamics predictor performs future prediction given candidate action sequences.

After the state estimator produces an estimated state $\hat{s}_T = \langle \hat{o}_{T}^{vis}, \xi_{T} \rangle$ from the $T$-step history, our dynamics model predicts into the future to evaluate potential action plans. 
The dynamics predictor $f$ is constructed similarly to the state estimator $g$, with two key differences: (i) it does not use tactile observations as input, and (ii) it is conditioned on frozen physics parameters estimated by $g$. Figure~\ref{fig:dynamics} illustrates this process.
The forward prediction happens recursively: For a step $t > T$, we construct a graph in the same way as in \secref{sec:estimator}, but excluding tactile observations from the particle attributes, \ie, $c^o_{i, t} = \xi_{t}$. Then, the dynamics predictor infers the particle positions at the next step $\hat{o}_{t+1}^{vis}$ as formulated in Equations~\ref{eqn:object_encode}-\ref{eqn:observation_predict}. The final state prediction is then $\hat{s}_{t+1}$ = $\langle \hat{o}_{t+1}^{vis}, \xi_{t} \rangle$. Note that the estimated physics parameters are not modified by the dynamics predictor. 

% \subsubsection{Training Procedure and Objective} \label{sec:loss}
\textbf{Training procedure and objective.} We train the state estimator and dynamics predictor jointly end-to-end on trajectories of sequential interaction data containing observations and robot actions.
For a training trajectory of length $H$, the state estimator estimates the first $T$ states, and the dynamics predictor predicts all remaining states. The estimation and prediction are all computed autoregressively. The loss is computed only on visual observations:
\begin{align}
    \L{} = \frac{1}{H} \sum_{t=0}^{H-1} || \hat{o}_{t}^{vis} - o_t^{vis} ||^2_2.
\end{align}

Previous works \citep{shi2022robocraft, shi2023robocraft, shi2023robocooka} use the earth mover's distance (EMD) or chamfer distance (CD) as the training loss, but these provide noisier gradients because EMD requires estimating point-to-point correspondence and CD is prone to outliers. Instead, we use mean squared error (MSE) 
% to train both our state estimator and dynamics predictor jointly end-to-end
as the objective, enabled by the point-to-point correspondences from our 3D point tracking~(\secref{sec:perception}).  \rebuttal{The details of the architecture and training procedure of the state estimator and dynamics predictor are in \appendref{appendix:dynamics_details}}. 
% Compared to EMD and CD, MSE provides a more accurate gradient for model optimization, because .

Note that the learning of the latent physics information is not explicitly supervised. The model is allowed to identify any latent parameters that enhance its ability to accurately estimate the current state and predict future outcomes. 
% This approach uncovers information that distinguishes between objects with varying physical characteristics. 
We provide an analysis on the learned physics parameters in \secref{sec:exp}.

\subsection{Model-Predictive Control}
\label{sec:control}
With the learned state estimator and dynamics predictor, we perform planning toward a particular goal by optimizing a cost function on predicted states over potential future actions. Concretely, we use Model Predictive Path Integral (MPPI) to perform this optimization~\citep{mppi2016}. 

Planning begins with sampling actions from an initial distribution performing forward prediction with the dynamics models. The cost is then computed on predicted states. Based on the estimated costs, we re-weight the action samples by importance sampling and update the distribution parameters. The process repeats for multiple interactions and we select the optimal execution plan.

For computational efficiency, we execute the first $K$ planning steps. While executing the actions, the robot records its tactile readings. After execution, it performs state estimation with the history of observations and re-plans for the next execution. \rebuttal{More implementation details on planning can be found in Appendix~\ref{appendix:planning}.}

\rebuttal{To summarize this section, a diagram of the entire system workflow including training and test-time deployment is available in Figure~\ref{fig:system}.}

%% file: text/040_setup.tex
\section{Experimental Setup}

\subsection{Physical Setup}
\begin{figure}[t]
    \centering
    \includegraphics[width=\columnwidth]{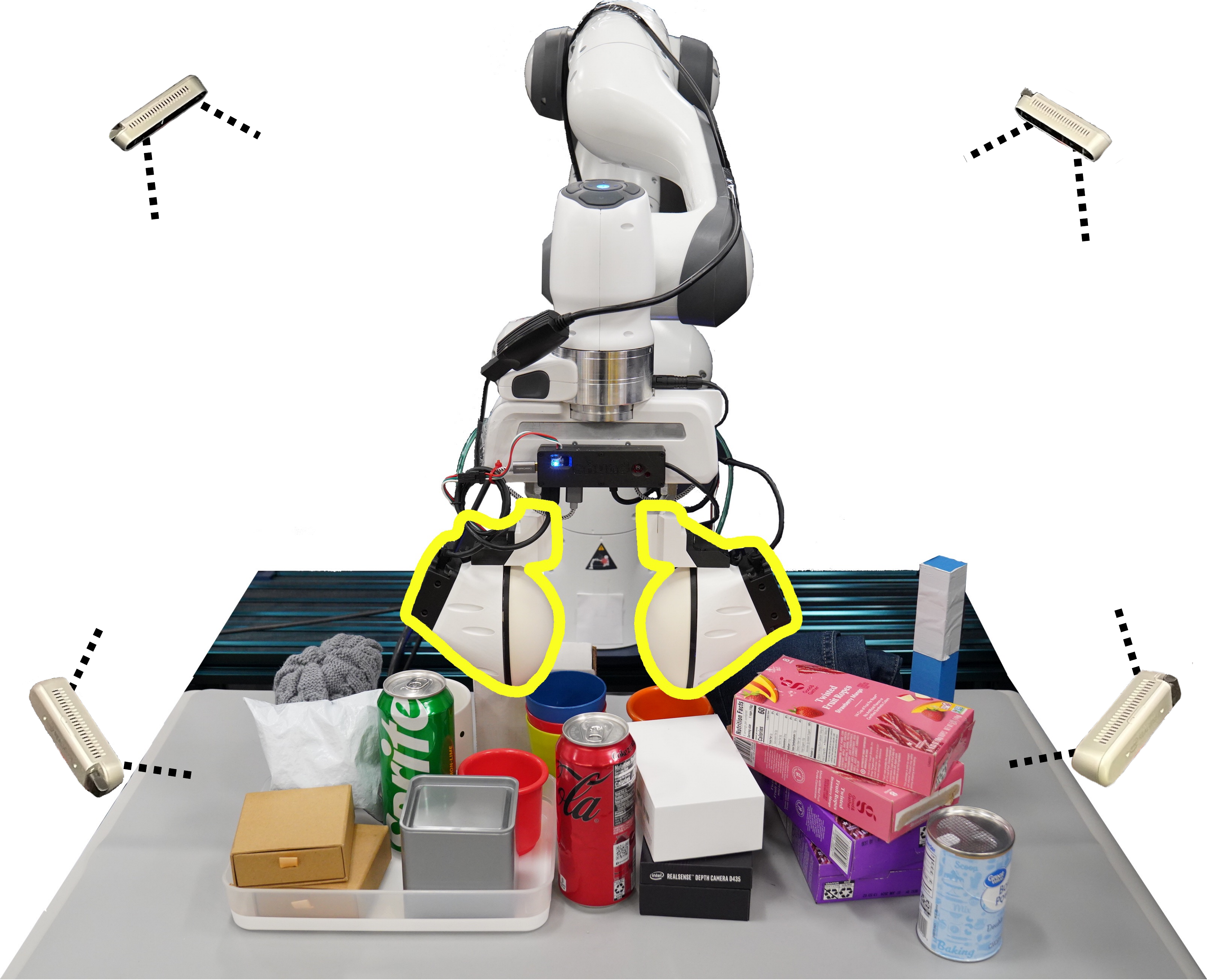} \vspace{-10pt}
    \caption{\textbf{Hardware overview}. Our experimental platform consists of a Franka Panda arm, two Soft-Bubble sensors, four RealSense D415 RGB-D cameras, and a diverse set of objects.}
    \vspace{-0.5cm}
    \label{fig:hardware}
\end{figure}

We set up our system on a Franka Emika Panda 7-DoF robotic arm. We use four Intel RealSense D415 cameras surrounding the robot and a pair of Soft-Bubble sensors for tactile feedback. We use 3D-printed connectors to attach the Soft-Bubble sensors to the robot. Each Soft-Bubble has a built-in RealSense D405 RGB-D camera. The RGB data are post-processed with an optical flow computation to approximate the force distribution over the bubble surface \citep{kuppuswamy2020softbubble}. 
% \boai{do we need to somehow point out this is done by its provided API instead of us? otherwise, they may ask why we do things in certain ways, and also just to make sure reviewers don't see this as our contribution}. 
Our hardware setup is depicted in Figure~\ref{fig:hardware}.

\subsection{Task Description}
We demonstrate our method on two tasks where the robot needs to handle objects with unknown physical properties and significant visual occlusion: manipulating a box with an in-hand tool and dense packing. 
 
\subsubsection{Non-Prehensile Box Pushing} This task focuses on manipulating rigid objects with varying mass distributions using an in-hand rod. The objective is to push a box to a goal pose with the minimum number of pushes. The robot has access to tactile feedback at all steps but only visual observations in between pushes, which corresponds to the real-world feedback loop frequency. The task is much more challenging than usual pushing tasks because (i) the boxes have different dynamics yet the same visual appearance; (ii) the robot has little visual feedback to identify box configurations; and (iii) the in-hand object can rotate and slip due to the highly compliant Soft-Bubble grippers. This is why we emphasize that our task is non-prehensile. This leads to rather complex physics interactions. To achieve effective planning, the robot needs to identify the box's properties from the tactile interaction history and adjust its predictions of the rod and box poses. 
% In essence, this task simulates the challenge faced by humans when, for example, manipulating objects on a table while engaging in conversation without their primary focus on the objects themselves.

We experiment with four boxes, each equipped with varying calibration weights attached to their inner layers to control their dynamics. We train our model on three of these boxes with identical visual appearances. During evaluation, we test our method on all four boxes including an additional one with a distinct visual appearance and mass distribution. 
% To keep the unseen box representative of real-world objects, we keep it intact as purchased. 

\subsubsection{Dense Packing} 

\begin{figure}[t]
    \centering
    \includegraphics[width=\columnwidth]{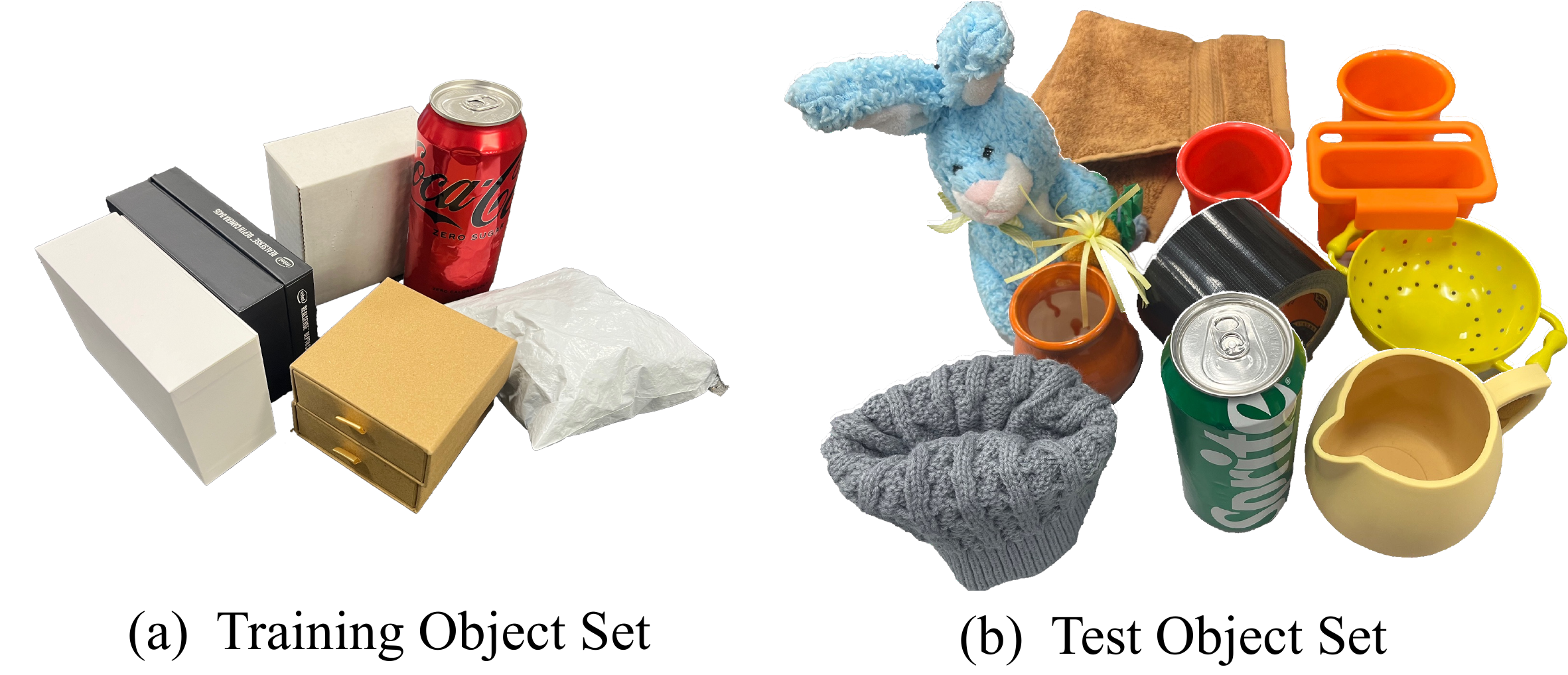} \caption{\textbf{Object sets for the packing task.} \rebuttal{The test objects are more complex than the training set visually, geometrically, and physically, to showcase the generalizability of our model.} }
    \vspace{-0.5cm}
    \label{fig:packing_object_sets}
\end{figure}

% This task focuses on enabling robots to interact with deformable objects. 
The goal of this task is to place an additional object in an already densely packed box. Due to heavy occlusions during task execution, the robot does not have access to meaningful visual feedback during robot execution other than the initial frame, but again tactile signals are always observed. To place the object into the occupied box, the robot needs to identify potentially deformable regions with tactile information and make space for the object via pushing actions. The robot needs to avoid inserting into infeasible regions to prevent hardware and object damage. We specify the box that contains the object as the goal and the robot can insert the object at any position as long as it fits inside. 

\rebuttal{To test the generalizability of learned models, we create train and test object sets (\figref{fig:packing_object_sets}). The test objects differ from the training objects in of visual appearance, surface geometry, and physical properties. During evaluation, we consider scenarios with only training objects and those with half or more of objects from the test set.}

% The goal is specified as a bounding box, which represents the entire region where the object can po be placed, \ie, the box to be packed. 

\subsection{Data Collection}
To generate diverse and safe interaction behaviors, we use human teleoperation for data collection. In the \textit{Non-Prehensile Box Pushing} task, for each weight configuration, we gather random interaction data for around 15 minutes. By ``random", we refer to the absence of labels typically present in demonstration data.
% We select one of three visually similar boxes and we control the mass distribution by attaching different calibration weights inside the box. 
%We teleoperate the robot using a 3Dconnexion SpaceMouse, guiding it to grasp a rod for interaction with the box. 
During these interactions, the end-effector approaches the box from various angles and contact locations, yielding diverse outcomes including translation and rotation, as well as relative movements between the in-hand object and the bubble gripper. The dataset contains approximately $12000$ total frames of interaction. 

For dense packing, we collect approximately $20$ minutes of teleoperated random interaction data with five unique objects, randomizing the initial configurations of the objects at the beginning of each interaction episode. Each episode includes various attempts at packing an object into the box and includes pushing and deforming objects, as well as in-hand slipping of the in-hand object in some trials. The dataset contains approximately $6000$ total frames of interaction.

% We train our models using the Adam optimizer \citep{adam}. It takes 25 and 8 hours to converge on the two tasks respectively with one NVIDIA RTX A5000 GPU. Details can be found in \appendref{appendix:training}. 

\subsection{Action Space}
 Though our dynamics model is orthogonal to the action space, suitable action abstractions are important for efficient planning and execution. 
\subsubsection{Non-Prehensile Box Pushing} To reduce the planning horizon and number of optimized parameters, we sample macro-actions during planning, which are defined as a linear push and represented by ${i, \theta, \alpha}$, where $i$ refers to the box particle index for end effector contact, $\theta$ denotes the angle of the push trajectory relative to the x-axis, and $\alpha$ represents the fraction of the distance covered before end effector-box contact along the entire push length. For dynamics prediction, the macro action is decomposed into smaller motions.
%\item Dense packing: As this involves a large state space, we use a hierarchical planning scheme to search for insertion actions efficiently. We first sample insertion points $\{x, y\}$ in the 2D plane at $z = Z$. Then, for each sample, we sample a sequence of correlated Gaussian noise parameterized by $\{\mu, \sigma\}$, which are the mean and the standard deviation of the distribution respectively, to represent the end-effector motion to execute. Such a combination of high-level motion and low-level fine-grained motion enables efficient and accurate planning and actuation.

\subsubsection{Dense Packing} As this task involves a large state space, we constrain the action space for planning efficiency. We first identify the outer objects in the box and compute feasible starting positions of actions nudging each object, determined by the geometric center of the object and its approximate radius. Then we sample straight-line push actions of varying lengths from each contact point towards the respective object centers. Similarly, the long push action is divided into small movements for dynamics prediction.

\subsection{Planning Cost Functions}
% We designed different planning loss functions for the two tasks. 
% \paragraph{ Box pushing: }
% We specify the goal state as a point cloud with point-to-point correspondence to the intermediate visual observations. Mean Squared Error is used to measure the distance between the current box particles and the target particles.

\subsubsection{Non-Prehensile Box Pushing} We specify the goal state as a point cloud and use MSE as the cost function. 
% It is transformed from the initial visual observation of the box to the target position and orientation. 

\subsubsection{Dense Packing} 
We specify a 2D goal region by uniformly sampling points in the area underneath the tray. We use a cost function that (i) penalizes the objects in the box from being pushed out of the boundary, (ii) encourages the robot to make space for placing the in-hand object by maximizing the 
% single-direction chamfer 
distance from target to object points, and (iii) rewards exploring different starting action positions. Mathematically, the loss function is 
% We specify the goal state as a set of points on a 2D region. We identify the box region by object segmentation, from which we uniformly sample points to represent the 2D area. Our loss function should achieve a few objectives: (i) preventing the objects in the box from being pushed out of the boundary by encouraging object points to be within the region, (ii) encouraging the robot to make space for placing the in-hand object, which can be achieved by maximizing the single-direction chamfer distance from target points to the object points, and (iii) encouraging the robot to switch to different action starting positions to explore pushing different objects in the box. Mathematically, the loss function is
\begin{align}
    \J(\hat{o}_t, o_{g}, a_t) = &  \sum_{x \in \hat{o}_t} \min_{y \in o_g} ||x - y||_2 - \sum_{y \in o_g} \min_{x \in \hat{o}_t} ||x - y||_2 \nonumber \\
    & + r * \mathds{1}_{[||a_{0, t}||_2 = 0]},
\end{align}
where $\hat{o}_t$ is the predicted object particles in the box, $o_g$ is the target point cloud, $||a_{0, t}||_2$ is the size of the first action, which is zero if it does not plan to switch to a different contact row, $r$ is a negative constant, and $\mathds{1}$ is an indicator function.

% \subsection{Evaluation Metrics}
% In the coming experiments section, we will investigate three questions with qualitative and quantitative results of both dynamics prediction and planning performance on the two robotic tasks. This subsection introduces the quantitative metrics we will be using for the purpose.

% For dynamics prediction, we use distance metrics consisting of MSE, EMD, and CD to provide a comprehensive evaluation of the prediction accuracy. For downstream planning, we have three metrics.
% \begin{enumerate}[label=\roman*.]
%     \item Success rate (SR): Success is defined as the cost of the final state lower than a certain threshold.
%     \item Final-state cost: This is the cost of the final state compared to the specified target.
%     \item Number of steps: One step is defined as one planning step. Having a smaller number of steps indicate higher efficiency of the policy. 
% \end{enumerate}

%% file: text/050_results.tex
\section{Experiments} \label{sec:exp}

In this section, we investigate the following questions. 
\begin{enumerate}[label=\roman*.]
    \item Does integrating tactile sensing information from prior interactions improve future prediction accuracy?  % To compare with \oursnotactile{}, \gnntactile{}, \gnnnotactile{} on offline dynamics prediction. 
    \item Do the latent representations learned by tactile dynamics models discover meaningful properties such as the physical properties of objects? % Clustering. Linear regression to predict COM/distribution of pressure (e.g., other ways to interpret the information in the physics parms with supervised learning)?
    \item Does our tactile-informed model-predictive control framework enable robots to solve tasks involving objects of unknown physical properties? 
    % \item Can the proposed tactile-informed dynamics model be used to tackle more complex contact-rich tasks?
\end{enumerate}

We first introduce our baselines and then present empirical results in the subsequent subsections. 
   
\subsection{Baselines and Prior Methods} \label{sec:baselines}

We compare our approach against three prior methods and baselines, including ablated versions of our model, previous work on dynamics learning, and a physics-based simulator: 
\begin{enumerate}[label=\roman*.]
    \item \textbf{\oursnotactile{}}: To study the effects of using tactile sensing in state estimation and dynamics prediction, we evaluate this ablation of our method, which zeroes out tactile input to the model. 
    \item \textbf{\gnntactile{}}: This approach differs from ours in that it treats the observations, \ie, visual and tactile observations $\langle o^{vis}, o^{tact} \rangle$, directly as the state representation, whereas \ours{} assumes partial observability of the underlying state and performs explicit state estimation. This can be viewed as an adaptation of previous work \citep{dpi, shi2022robocraft, shi2023robocraft, shi2023robocooka} to include an additional tactile observation component. With this baseline, we seek to study different state representations and our strategy of separating state estimation from dynamics prediction. 
    \item \textbf{\simulator{}}: We also compare our method to using a physics-based simulator for dynamics prediction after performing system identification of explicit physical parameters. We use heuristics to convert observed point clouds into body positions and orientations in the 2D physics simulator Pymunk~\citep{pymunk}. For system identification, we estimate the mass, center of gravity, and friction parameters from the initial and current visual observations with covariance matrix adaptation~\citep{hansen2016cma}.
\end{enumerate}
% Here we note that, our baselines can be viewed as offline model-based RL methods (\secref{sec:related-rl}), which can help us calibrate the performance of \ours{} in the context of RL literature. Specifically, \oursnotactile{} can be viewed as an RL method that only uses sparse visual feedback for policy learning, and \gnntactile{} is an RL method that naively learns a policy in the raw visual and tactile observation space without accounting for the non-Markovian nature of the task. As we will show later, missing sensing modality or treating the task as fully observable MDP can lead to much poorer robotic performance in the real world. 

\rebuttal{The considered methods, including our approach, share some conceptual components with prior offline model-based reinforcement learning (RL) methods (\secref{sec:related-rl}), although with very different concrete instantiations. Each method either learns the full environment dynamics, or in the case of \textit{\simulator{}}, performs system identification from a static dataset. All compared methods use the dynamics models to perform model-predictive control via sampling-based planning. Specifically, \textit{\oursnotactile{}} can be framed as a model-based RL method (\eg, \citep{daydreamer, world-models, visual-foresight}) that uses only sparse visual observations for model learning. On the other hand, \textit{\gnntactile{}} treats visual and tactile observations as the state, overlooking the partially observable nature of the task. Our upcoming results demonstrate that our integration of multi-modal perception and physical parameter estimation leads to superior performance in challenging task domains.}

\subsection{Evaluating Dynamics Prediction} \label{sec:dynamics_prediction}

% \begin{table*}[t]
% \centering
% \begin{tabular}{lccc}
% \toprule
% Method & MSE $\downarrow$ & EMD $\downarrow$ & CD $\downarrow$ \\
% \midrule
% \textbf{\ours{}} &  \textbf{0.00148 $\pm$ 0.00014} & \textbf{0.02974 $\pm$ 0.00138} & \textbf{0.03463 $\pm$ 0.00130} \\
% \oursnotactile{} & 0.00175 $\pm$ 0.00015 & 0.03341 $\pm$ 0.00145 & 0.03796 $\pm$ 0.00134 \\
% \bottomrule
% \end{tabular}
% \caption{50-Step dynamics prediction results on the dense packing dataset. 95\% confidence interval reported.}
% \label{tab:dynamics_quantitative_packing}
% \end{table*}

% \begin{table*}[t]
% \caption{50-Step Dynamics Prediction Results}
% \centering
% \begin{tabular}{lccccccccc}
% \toprule
% Method & \multicolumn{3}{c}{MSE $\downarrow$} & \multicolumn{3}{c}{EMD $\downarrow$} & \multicolumn{3}{c}{CD $\downarrow$} \\
% \cmidrule(lr){2-4} \cmidrule(lr){5-7} \cmidrule(lr){8-10} 
%  & 25\% & Median & 75\% &  25\% & Median & 75\% & 25\% & Median & 75\% \\
% \midrule
% \ours{}&  0.00010 & 0.00036 & 0.00147 & 0.01154 & 0.01995 & 0.03604 & 0.02053 & 0.02611 & 0.03683 \\
% \oursnotactile{} & 0.00016 & 0.00056 & 0.00198 & 0.01440 & 0.02253 & 0.04091 & 0.02271 & 0.02862 & 0.04116  \\
% \gnntactile{} & 0.00024 & 0.00062 & 0.00258 & 0.02158 & 0.03041 & 0.05340 & 0.03491 & 0.04235 & 0.05863 \\
% \gnnnotactile{} & 0.00018 & 0.00062 & 0.00242 & 0.01956 & 0.02941 & 0.05238 & 0.03251 & 0.04002 & 0.05730 \\
% \bottomrule
% \end{tabular}
% \label{tab:dynamics_quantitative}
% \end{table*}

\begin{table}[t]
\centering
\vspace{0.2cm}
% \resizebox{\linewidth}{!}{
% \footnotesize
\setlength{\tabcolsep}{0.8mm}{
\begin{tabular}{clccc}
\toprule
Task & Method & MSE *1e-3 $\downarrow$ & EMD *1e-2 $\downarrow$ & CD *1e-2 $\downarrow$ \\
% Task & Method & MSE  ($\times 10^{-3}$) $\downarrow$ & EMD ($\times 10^{-2}$) $\downarrow$ & CD ($\times 10^{-2}$) $\downarrow$ \\
\midrule

& \textbf{\ours{}} &  \textbf{1.48 $\pm$ 0.14} & \textbf{2.97 $\pm$ 0.14} & \textbf{3.46 $\pm$ 0.13} \\
Box& \oursnotactile{} & 1.75 $\pm$ 0.15 & 3.34 $\pm$ 0.15 & 3.80 $\pm$ 0.13 \\
Pushing& \gnntactile{} & 2.11 $\pm$ 0.17 & 4.32 $\pm$ 0.16 & 5.40 $\pm$ 0.16 \\
% & \gnnnotactile{} & 2.05 $\pm$ 0.16  & 4.17 $\pm$ 0.16 & 5.17 $\pm$ 0.16 \\
& Physics-based sim. & 2.65 $\pm$ 0.18 & 4.11 $\pm$ 0.17 & 4.57 $\pm$ 0.16 \\
\midrule
Dense & \textbf{\ours{}} & \textbf{0.070 $\pm$ 0.005} & \textbf{1.12 $\pm$ 0.036} & \textbf{2.01 $\pm$ 0.050} \\
Packing& \oursnotactile{} & 0.088 $\pm$ 0.006 & 1.18 $\pm$ 0.043 & 2.04 $\pm$ 0.058 \\

% unscaled below
% \multirow{5}{*}{Box Pushing} & \textbf{\ours{}} &  \textbf{0.00148 $\pm$ 0.00014} & \textbf{0.02974 $\pm$ 0.00138} & \textbf{0.03463 $\pm$ 0.00130} \\
% & \oursnotactile{} & 0.00175 $\pm$ 0.00015 & 0.03341 $\pm$ 0.00145 & 0.03796 $\pm$ 0.00134 \\
% & \gnntactile{} & 0.00211 $\pm$ 0.00017 & 0.04318 $\pm$ 0.00161 & 0.05396 $\pm$ 0.00158 \\
% & \gnnnotactile{} & 0.00205 $\pm$ 0.00016  & 0.04166 $\pm$ 0.00160 & 0.05170 $\pm$ 0.00155 \\
% & \simulator{}  & 0.00265 $\pm$ 0.00018 & 0.04111 $\pm$ 0.00169 & 0.04574 $\pm$ 0.00164 \\
% \midrule
% \multirow{2}{*}{Dense Packing} & \textbf{\ours{}} & \textbf{0.000070 $\pm$ 0.000005} & \textbf{0.011180 $\pm$ 0.000355} & \textbf{0.020086 $\pm$ 0.000499} \\
% & \oursnotactile{} & 0.000088 $\pm$ 0.000006 & 0.011768 $\pm$ 0.000425 & 0.020424 $\pm$ 0.000580 \\
\bottomrule
\end{tabular}}
\caption{\textbf{Long-horizon dynamics prediction results on the two task datasets.} Errors represent a 95\% confidence interval.}
\label{tab:dynamics_quantitative_pushing}
\vspace{-0.5cm}
\end{table}

Results are summarized in \tabref{tab:dynamics_quantitative_pushing}. On the \textit{Non-Prehensile Box Pushing} task, \ours is significantly better than alternative methods in all metrics. Compared to \oursnotactile, \ours can better estimate the mass distribution of the boxes, which is crucial in predicting the translation and rotation accurately. In contrast, when using tactile and visual observations directly as the state representation (\gnntactile), the performance is even worse than RoboPack without tactile information. We hypothesize that this is because the model has very high errors in learning to predict future tactile readings because of the intricate local interactions between the Soft-Bubble grippers and the object. 
% For example, as the soft bubble surface is compliant, the in-hand object can rotate over time, causing the bubble surface to deform and eventually leading to noisy force readings.
The difficulty in learning to predict tactile reading may distract the model from learning to predict visual observations accurately.
% , causing the performance to be worse than our non-tactile model baseline. 

\begin{figure}[t!]
    \centering
    \includegraphics[width=\columnwidth]{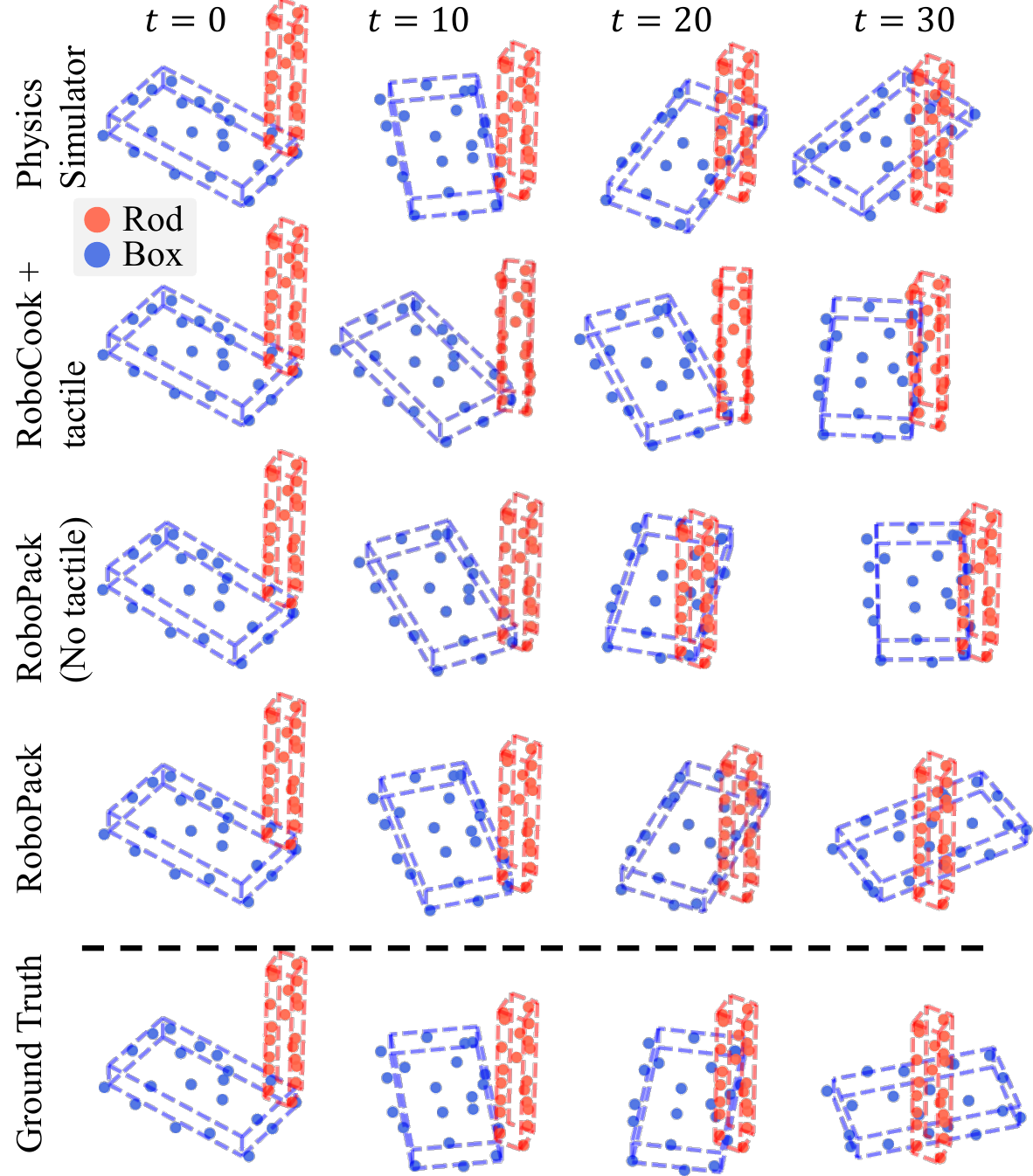} \caption{ \textbf{Qualitative results on dynamics prediction.} Predictions made by our model compared to baseline methods in the \textit{Non-prehensile Box Pushing} task. Red dots indicate the rod and blue dots represent the box. Our method closely approximates the ground truth and outperforms all the baseline methods. For visualization, the blue dashed lines outline box contours and red dashed lines show in-hand object contours.}
    \vspace{-0.5cm}
    \label{fig:dynamics_qualitative}
\end{figure}

Comparing \ours to a physics-based simulator baseline, we find that the simulator performs poorly on dynamics prediction for a few potential reasons, including (i) limited visual feedback for performing system ID, and (ii) the simulator's parameter space may not capture the full range of real-world dynamics given the complex interactions between the compliant bubble and in-hand tool and rotating tool and the box. To illustrate the difference in model predictions, qualitative results are presented in \figref{fig:dynamics_qualitative}.

For the \textit{Dense Packing} task, our model outperforms the best baseline on the pushing task, \oursnotactile{}. \rebuttal{We note that in this task, object movements are minimal and object deformation is the major source of particle motions. Metrics such as EMD and CD that emphasize global shape and distribution but are insensitive to subtle positional changes cannot differentiate the two methods in a statistically significant way. However, for the MSE loss, which measures prediction error for every point, \ours{} is significantly better than the baseline, indicating its ability to capture fine details of object deformation. This subtle performance difference between the two methods in dynamics prediction turns out to have a significant effect on real-world planning (\secref{sec:robotic-task}).}

% Since most objects remain static in this task, loss scales are smaller than in the pushing task.
% We found that the baseline learns short-cut visual cues from the initial visual frame, \eg, inferring that an object will not move based on the inhand object pose. However, during deployment, this fails because of the longer task horizon and lack of initial visual cues (see \secref{sec:robotic-task}).

\begin{figure*}[t]
    \centering
    \vspace{-0.3em}
    \includegraphics[width=1.9\columnwidth]{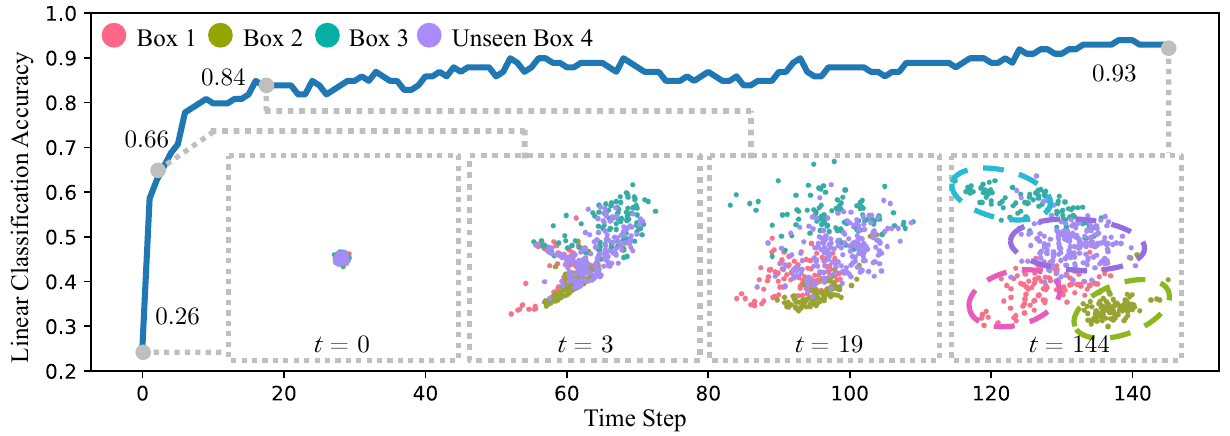} 
    \caption{\textbf{Analysis of learned physics parameters.} We assess our state estimator across 145-step trajectories and record the estimated physics parameters at each step. PCA visualizations at four distinct timesteps show that the physics parameters gradually form clusters by box type. We also employ a linear classifier trained on these parameters to accurately predict box types to demonstrate these clusters' linear separability. The classifier's improving accuracy across timesteps underscores the state estimator's proficiency in extracting and integrating box-specific information from the tactile observation history. 
    % Notably, during training, the state estimator has only seen trajectories with a maximum length of 25 steps, but it can generate meaningful physics parameters for 145-step trajectories. This highlights the generalization capability of our state estimator.
    }
    % \vspace{-0.1cm}
    \label{fig:physics_params}
\end{figure*}

\begin{table*}[t]
\centering
\begin{tabular}{llccccc}
% \toprule
% For reference, this is the map of table: our usual box convention:
% Box1 = Box3 
% Box2 = Box7
% Box3 = Box5
% Box4 = Box6 (unseen)

& & \includegraphics[height=2.5em]{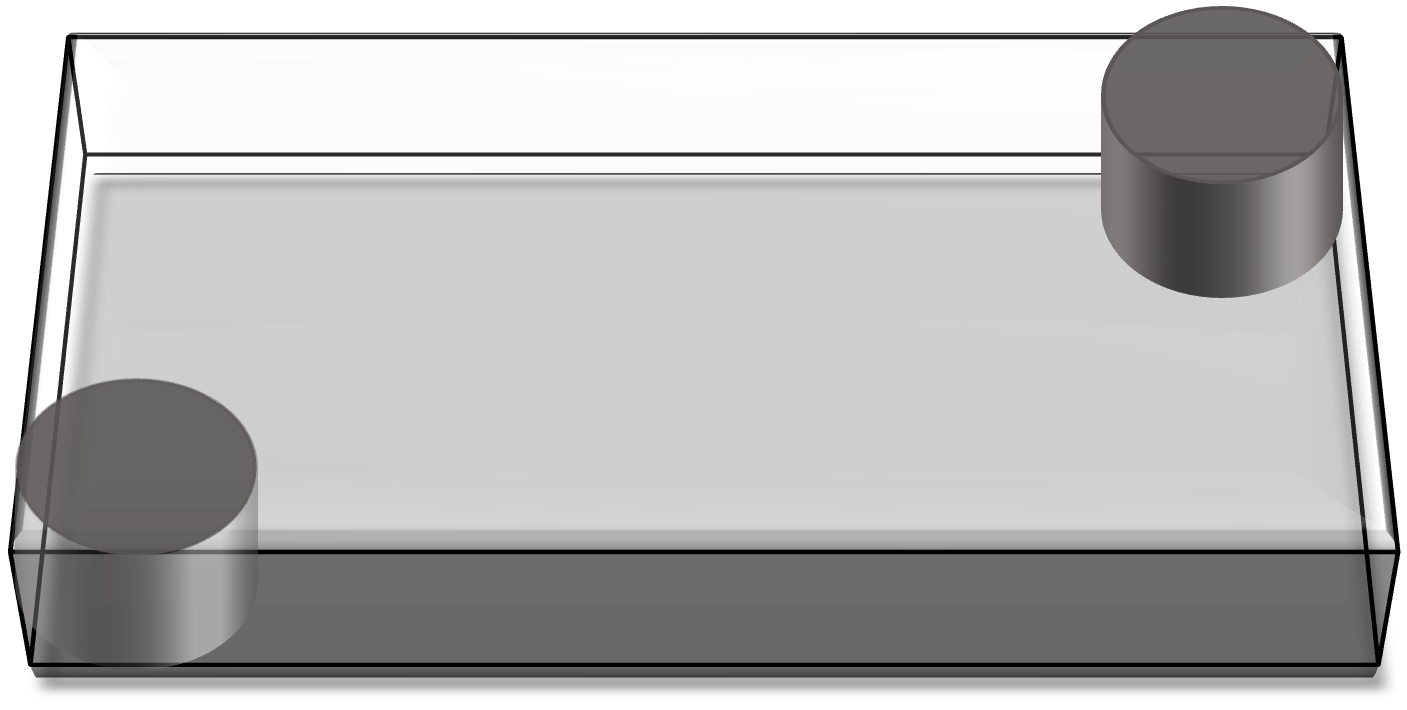} & \includegraphics[height=2.5em]{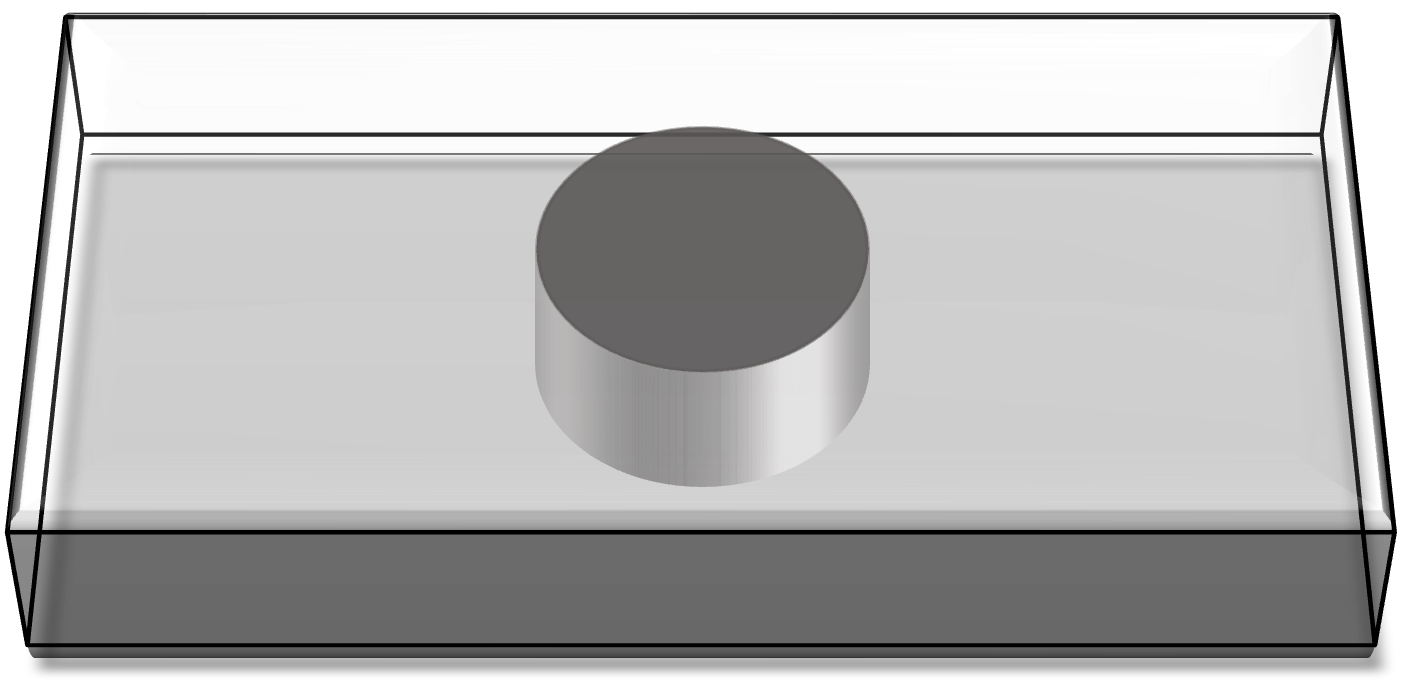} & \includegraphics[height=2.5em]{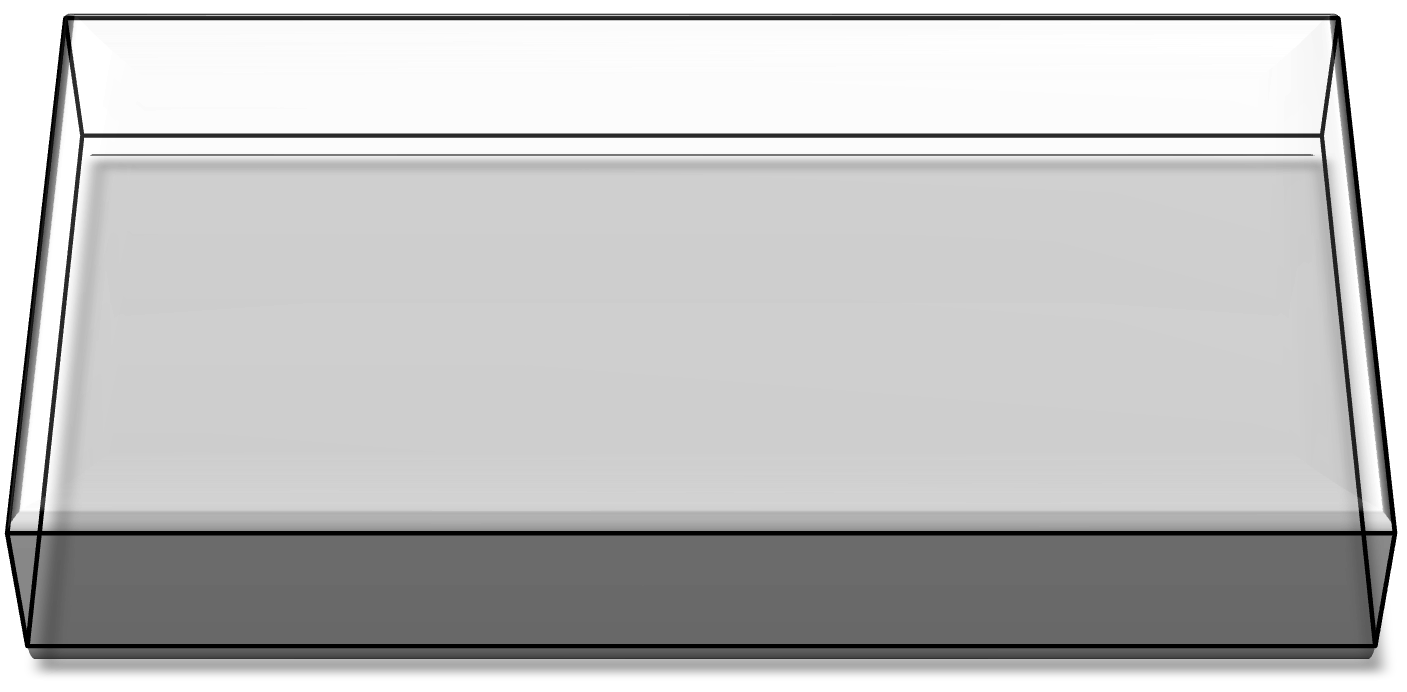} & \includegraphics[height=2.5em]{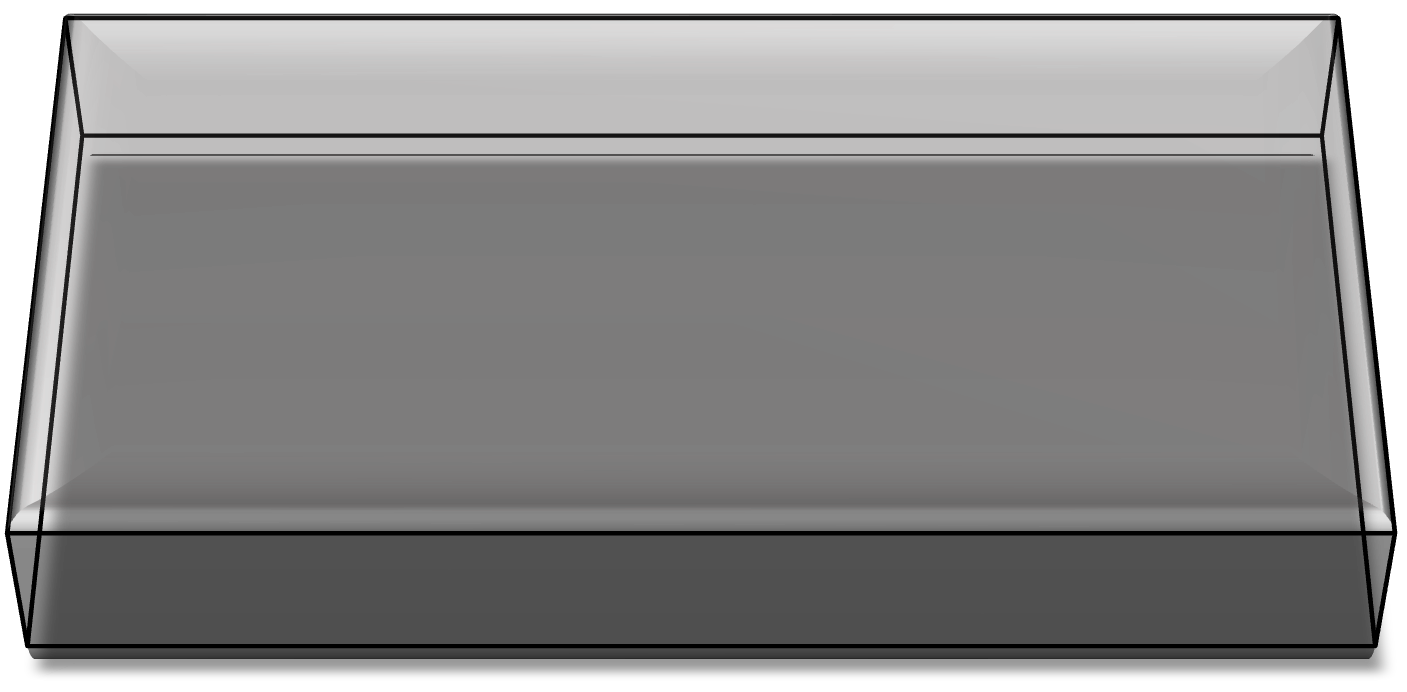} &\\

Method & Metric & Box 1 & Box 2 & Box 3 & Box 4 (unseen) & Aggregated \\

\midrule
\multirow{3}{*}{\ours{}} & \MSE{} $\downarrow$ & 0.0164 $\pm$ 0.004 & 0.0165 $\pm$ 0.004 & 0.0137 $\pm$ 0.003 & 0.0156 $\pm$ 0.001 & \textbf{0.0156 $\pm$ 0.002} \\
& \pushes{} $\downarrow$ & 5.0 $\pm$ 1.20 & 5.40 $\pm$ 1.49 & 4.8 $\pm$ 1.24 & 6.0 $\pm$ 1.10 & \textbf{5.3 $\pm$ 0.64} \\
& \SR{} $\uparrow$ & 4 / 5 & 4 / 5 & 4 / 5 & 4 / 5 & \textbf{16 / 20}  \\

\midrule
\multirow{3}{*}{\oursnotactile{}} & \MSE{} $\downarrow$ & 0.0612 $\pm$ 0.027 & 0.0141 $\pm$ 0.003 & 0.0250 $\pm$ 0.001 & 0.0264 $\pm$ 0.005 & 0.0317 $\pm$ 0.008 \\
& \pushes{} $\downarrow$ & 8.2 $\pm$ 0.99 & 5.0 $\pm$ 2.82 & 10.0 $\pm$ 0 & 8.2 $\pm$ 1.07 & 7.85 $\pm$ 0.63  \\
& \SR{} $\uparrow$ & 2 / 5 & 4 / 5 & 0 / 5 & 2 / 5 & 8 / 20 \\

\midrule
\multirow{3}{*}{\gnntactile{}} & \MSE{} $\downarrow$ & 0.0459 $\pm$ 0.018 & 0.0607 $\pm$ 0.022 & 0.0418 $\pm$ 0.009 & 0.0438 $\pm$ 0.017  & 0.0480 $\pm$ 0.009 \\
& \pushes{} $\downarrow$ & 8.2 $\pm$ 1.21 & 7.4 $\pm$ 1.73 & 9.2 $\pm$ 0.72 & 8.8 $\pm$ 1.07 & 8.4 $\pm$ 0.64  \\
& \SR{} $\uparrow$ & 2 / 5 & 2 / 5 & 1 / 5 & 1 / 5  & 6 / 20   \\

\midrule
\multirow{3}{*}{\simulator{}} & \MSE{} $\downarrow$ & 0.0237 $\pm$ 0.004 & 0.0184 $\pm$ 0.003 & 0.0273 $\pm$ 0.012  & 0.0220 $\pm$ 0.004 & 0.0230 $\pm$ 0.003 \\
& \pushes{} $\downarrow$ & 8.4 $\pm$ 0.92 & 6.0 $\pm$ 0.18 & 7.4 $\pm$ 1.19 & 7.4 $\pm$ 1.49 & 7.3 $\pm$ 0.71 \\
& \SR{} $\uparrow$ & 2 / 5 & 3 / 5 & 3 / 5 & 2 / 5  & 10 / 20  \\
% \midrule
% \multirow{3}{*}{\gnnnotactile{}} & \MSE{} & 0.0000 $\pm$ 0.0000 & 0.0000 $\pm$ 0.0000 & 0.0000 $\pm$ 0.0000  & 0.0000 $\pm$ 0.0000 & 0.0000 $\pm$ 0.0000 \\
% & \pushes{} & 0.0000 $\pm$ 0.0000 & 0.0000 $\pm$ 0.0000 & 0.0000 $\pm$ 0.0000 & 0.0000 $\pm$ 0.0000 & 0.0000 $\pm$ 0.0000 \\
% & \SR{} & 0 / 5 & 0 / 5 & 0 / 5 & 0 / 5  & 0 / 20  \\

\bottomrule
\end{tabular}

\vspace{0.1em}
\caption{\textbf{Per-configuration results on the non-prehensile box pushing task}. 
We report the minimum error to goal across $10$ plan executions per trial, trial success rates, and number of execution steps to solve the task. 
A trial is labeled as a success if it achieves an error lower than $0.02$ for point-wise MSE within 10 pushes.
% , and the number of pushes required to reach the threshold (we label the max $10$ for failures). We use a success threshold of $0.02$ for point-wise MSE, roughly corresponding to a Euclidean distance of $2$cm.
}
\vspace{-0.3cm}

\label{tab:pushing}
\end{table*}

\begin{figure*}[t]
    \centering
    \includegraphics[width=\linewidth]{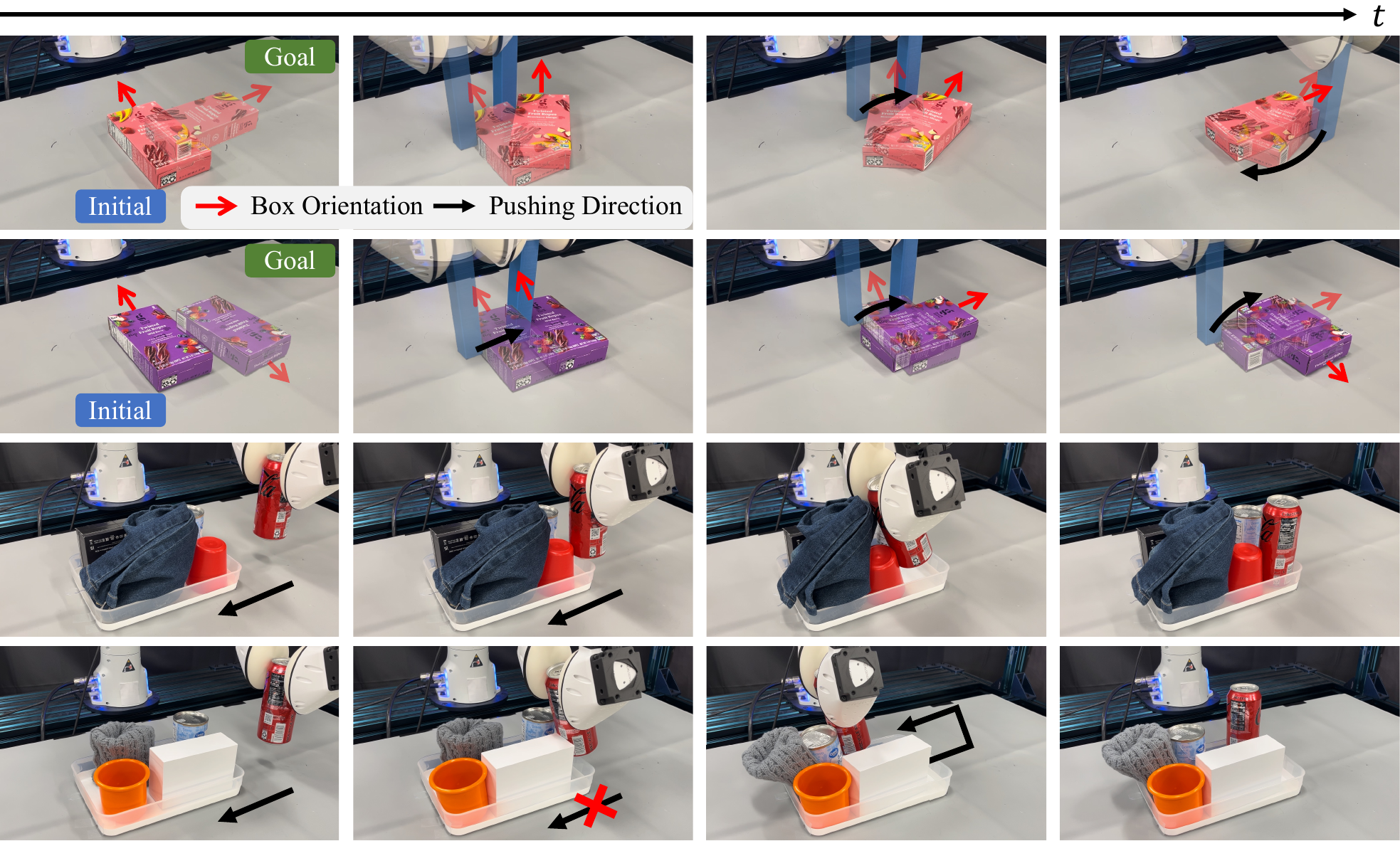} 
    \caption{\textbf{Non-prehensile box pushing and dense packing.} In the \textit{Non-prehensile Box Pushing} task, we demonstrate that our robot can push a box with unknown mass distribution from a starting pose to a target pose. 
    % Note that our box pushing is non-prehensile because the in-hand object is not fixed. 
    We show that our method can generalize to unseen targets and box configurations in the first two rows. In the \textit{Dense Packing} task, we demonstrate that \ours effectively identifies feasible insertion rows in a tray, minimizing excessive force to prevent hardware damage for incorrect contact locations while taking pushing actions decisively at correct contact points for efficient task completion. The last two rows illustrate that our method can adapt to objects with different visual appearances, shapes, and deformability.}
    \vspace{-0.2cm}
    \label{fig:robotic_qualitative}
\end{figure*}

\begin{figure}[t!]
    \centering
    \vspace{-10pt}
    \includegraphics[width=0.97\columnwidth]{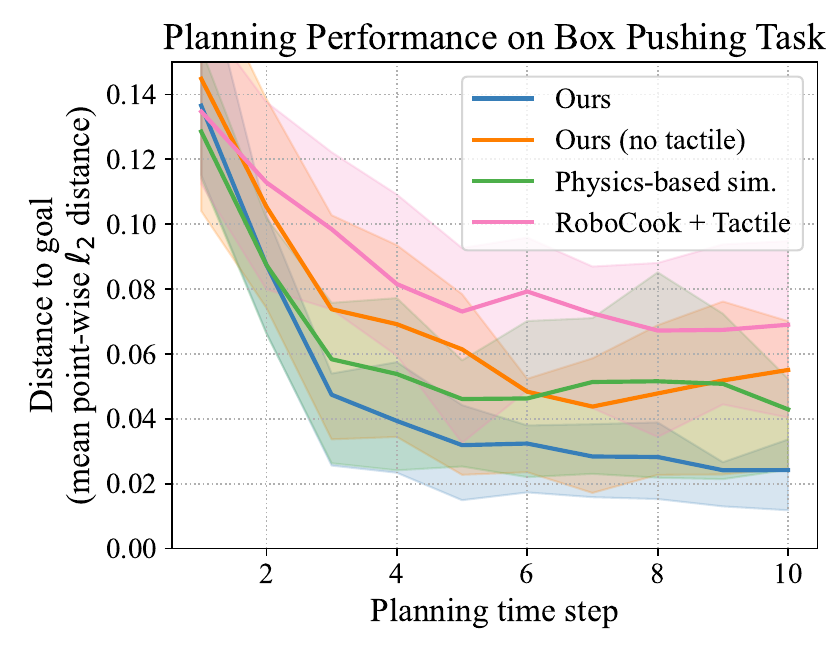} 
    \vspace{-0.2cm}
    \caption{\textbf{Real-world planning performance on the box pushing task.} Shaded regions denote the first and third quantiles. Note that different methods generally perform well on easier cases, leading to overlap between shadow regions. Our method has stable performance even for hard ones: its 75-percentile error is lower than the mean error of all other methods.}
    \vspace{-0.5cm}
    \label{fig:boxpushing-results}
\end{figure}

\subsection{Analysis of Learned Physics Parameters}

% \begin{figure}
%     \centering
%     \includegraphics[width=\columnwidth]{experiment/params_acc.png} 
%     % \vspace{-5pt}
%     \caption{Plot of classification accuracy of box types based on estimated physics parameters at a given history length. As more history information becomes available, the physics parameters enable more accurate classification as more history information is available. }
%     \label{fig:classification_acc}
% \end{figure}

% \begin{figure}[t]
%     \centering
%     % \setlength{\tabcolsep}{2pt} 
%     % \renewcommand{\arraystretch}{2} 
%     % \vspace{15pt}
%     \begin{tabular}{cc}
%          \includegraphics[width=0.45\columnwidth]{experiment/pca/pca_plot_0.png} &
%          \includegraphics[width=0.45\columnwidth]{experiment/pca/pca_plot_1.png} \\
%          t = 0 & t = 2 \\
%          \includegraphics[width=0.45\columnwidth]{experiment/pca/pca_plot_19.png}  &
%          \includegraphics[width=0.45\columnwidth]{experiment/pca/pca_plot_149.png}  \\
%         t = 19 & t = 149 \\
%     \end{tabular}
%     \caption{Visualizations of learned physics parameters. Each color denotes one box type. It can be observed that, as the model integrates more history information into the state representation, the physics parameter becomes more indicative of the specific box type. }
%     \label{fig:physics_params_pca}
% \end{figure}

In this subsection, we seek to provide some quantitative and qualitative analyses of the latent representation learned by the state estimator. As it gives more direct control of object properties, we use our dataset collected for the \textit{Non-Prehensile Box Pushing} task for the analysis. 

To understand if the representation contains information about box types, we first attempt to train a linear classifier to test if there the features learned for different boxes are linearly separable in the latent space. We test the state estimator on $145$-step trajectories in the testing data, which typically involves three to five pushes on the box. 
% The physical interactions between the box and the in-hand object may reveal information about the physical properties of the box. 
The classification accuracy of physics parameters $\xi_t$ as more and more interaction information is processed is shown in \figref{fig:physics_params}. It can be observed that as history information accumulates, the latent physics vectors become more indicative of the box type. In particular, the state estimator can extract considerable information in the first 20 steps, which is approximately the average number of steps it takes to complete an initial push. Furthermore, note that the state estimator only observes a history of no more than 25 steps during training, but it can generalize to sequences four times longer in this case. 

To qualitatively inspect the learned representations, we perform principal component analysis, reducing the learned latent vectors from $\mathcal{R}^{16}$ to $\mathcal{R}^2$. \figref{fig:physics_params} shows the low-dimensional embeddings as the number of interaction time steps incorporated into the latents grows. 
% The physics parameters are initialized as Gaussian noise at $t=0$. 
We can see that as time progresses, the estimated latents become increasingly separated into clusters based on the physical properties (\ie, mass distributions in this case) of the manipulated object. The separation increases the most between $t=1$ and $t=20$, which is consistent with our observation in \figref{fig:physics_params} that longer histories than a certain threshold yield marginal returns.

Collectively, the results indicate that our state estimator indeed learns information related to physical properties based on interaction histories.

\subsection{Benchmarking Real-World Planning Performance} \label{sec:robotic-task}

% \begin{table}[t]
% \centering
% \normalsize
% \begin{tabular}{lcc}
% Method & Four objs. & Five unseen objs.\\
% \midrule
% \ours{} & $\mathbf{80}$\% & $\mathbf{60}\%$ \\
% \oursnotactile{} & $40$\% & $50$\% 
% \end{tabular}
% \label{tab:packing}
% \caption{Success rates on dense packing task. Success rates are computed across $15$ and $10$ scenarios for the four and five unseen object categories respectively.}
% \end{table}

Next, we evaluate the performance of our approach in solving real-world robotic planning tasks. 

% Recall that the box-pushing task requires the robot to use a rod as a tool to manipulate a box of varying mass distribution to a particular position and orientation. For all learned methods, we include interaction data with three box configurations in the training set, and hold out one configuration as an unseen test scenario. 

For \textit{Non-Prehensile Box Pushing}, we present quantitative results in Figure~\ref{fig:boxpushing-results} and Table~\ref{tab:pushing}. We can see that our method both achieves lower final error as measured by point-wise MSE (\tabref{tab:pushing}) and makes progress toward goals more quickly (\figref{fig:boxpushing-results}) than other methods. The gap in performance between our model and \oursnotactile demonstrates the benefits of using tactile sensing in this task. While the physics-based simulator achieves the strongest performance of the baselines, \rebuttal{it is not able to achieve as precise control as our method, taking more pushes to finish the task yet ending with higher MSE loss.} We hypothesize this is because it can only infer dynamics of limited complexity via properties such as friction or mass center/moment. It also requires significant manual designs to construct the simulation for each task. Finally, \gnntactile has the poorest control performance, consistent with its high dynamics prediction error on the test set. We hypothesize that the poor performance of this method is due to the difficulty of learning to predict future tactile observations, which are high-dimensional and sensitive to precise contact details.

% On the \textit{Dense Packing} task, we would like to compare against the best method on object pushing, \ie, the physics-based simulator. However, it is not feasible to obtain corresponding object models for the diverse and complex objects in this task, so we use \oursnotactile as the baseline, which has similar performance to the physics-based method. We test both methods on packing in $15$ object configurations and they achieve success rates of $80\%$ and $40\%$ respectively. This shows that our method is much more effective in identifying objects that are deformable or pushable, which consequently enables the robot to insert the object at feasible locations. Examples of our experiments are illustrated in \figref{fig:robotic_qualitative}. We show that, despite our method having only seen rectangular boxes and plastic bags in the training set, it can generalize to objects with different visual appearances, geometries, and physical properties, such as the plastic cups, cloth, and yarn in the examples.

\begin{table}[t]
\centering
\normalsize
\begin{tabular}{lcc}
Method & Seen Objects & Unseen Objects \\
\midrule
\ours{} & $\mathbf{12/15}$ & $\mathbf{10/15}$ \\
\oursnotactile{} & $6/15$ & $5/15$ 
\end{tabular}
\caption{\rebuttal{\textbf{Success rates on the dense packing task.} In the Unseen Objects setting, half or more of the objects in the tray are unseen. A trial is considered successful if the robot correctly determines feasible insertion locations and creates enough space (through deformation) to pack the object. The robot automatically attempts to pack the object when its end effector $y$-position exceeds a given threshold.}} 
\label{tab:packing}
\vspace{-0.3cm}
\end{table}

\rebuttal{For the \textit{Dense Packing} task, we would ideally compare our method against the baseline with the best results on non-prehensile box pushing: the physics-based simulator. However, this is impractical for this task, because it is infeasible to obtain corresponding object models for the diverse and complex objects in this task or to estimate objects' explicit physics parameters without visual feedback. Thus, we compare against the best among the remaining baselines instead, \ie, \oursnotactile{}. We test on scenarios containing only training objects (\textit{Seen Objects}) as well as scenarios where half or more of the objects are from the test set (\textit{Unseen Objects}). Results on both settings, shown in \tabref{tab:packing}, indicate that our method is more effective in identifying objects that are deformable or pushable, which consequently enables the robot to insert the object at feasible locations. Examples of our experiments are illustrated in \figref{fig:robotic_qualitative}. Despite our method having only seen rectangular boxes and plastic bags in the training set, it can generalize to objects with different visual appearances, geometries, and physical properties, such as the cups, cloth, and hat in the examples.}

%% file: text/060_conclusion.tex
\section{Discussion} \label{sec:conclusion}
We presented \ours{}, a framework for learning tactile-informed dynamics models for manipulating objects in multi-object scenes with varied physical properties. By integrating information from prior interactions from a compliant visual tactile sensor, our method adaptively updates estimated latent physics parameters, resulting in improved physical prediction and downstream planning performance on two challenging manipulation tasks, \textit{Non-Prehensile Box Pushing} and \textit{Dense Packing}. We hope that this is a step towards robots that can seamlessly integrate information with multiple modalities from their environments to guide their decision-making.

\rebuttal{In this paper we demonstrated our approach on two specific tasks, but our framework is generally applicable to robotic manipulation tasks using visual and tactile perception. To extend it to other tasks, one needs to adapt the cost function and planning module to the task setup, but the perception, state estimation, and dynamics prediction components are general and task-agnostic. 
For future work, we seek to develop dynamics models that can efficiently process higher-fidelity particles to model fine-grained object deformations. Integrating alternative trajectory optimization methods with our learned differentiable neural dynamics models is another promising direction. Finally, incorporating additional physics priors into the dynamics model could further improve generalization. }

% \clearpage

%% file: text/070_appendix.tex
\clearpage
\appendices

\section{Model Architecture and Training} % \label{appendix:training}
\subsection{\rebuttal{Tactile Autoencoder}} \label{appendix:ae_details}
\rebuttal{Both the encoder and decoder are two-layer MLPs with hidden dimension 32 and ReLU activations. The encoder maps the raw point-wise tactile signal to latent space, then the decoder maps it back to the original dimension. The autoencoder is trained with MSE loss using the following hyper-parameters:}

\begin{table}[htb]
    \centering
    \begin{tabular}{lr}
        Hyperparameter & Value  \\
        \toprule
        Learning rate &  5e-4 \\
        Optimizer & Adam~\citep{adam} \\
        Batch size & 32 \\
        Latent space dimension & 5 \\ 
    \end{tabular}
    \caption{\textbf{\rebuttal{Hyperparameters for auto-encoder training.}}}
    \label{tab:ae_training_details}
\end{table}

\subsection{State Estimator and Dynamics Predictor} \label{appendix:dynamics_details}
We use the same hyperparameters to train dynamics models for the nonprehensile box pushing and dense packing tasks, which are shown in Table~\ref{tab:model_training_details} For graph construction, we connect any points within a radius of $0.15$. We train the state estimator and dynamics model jointly, using sequences of length $25$. To prevent the model from overfitting to a specific history length, which could vary at deployment time, we use the first $k$ steps in a sequence as the history, $k \sim \text{Uniform}(0, 24)$. To stabilize training, we restrict the magnitude of the rotation component of predicted rigid transformations for a single step to be at most $30$ degrees, which is much larger than any rotation that occurs in our datasets. Model training converges within 25 and 8 hours on the two tasks respectively with one NVIDIA RTX A5000 GPU.

For baselines \oursnotactile{} and \gnntactile{}, we performed a hyper-parameter sweep and the optimal training parameters are the same as \ours{} described above.

\begin{table}[htb]
    \centering
    \begin{tabular}{lr}
        Hyperparameter & Value  \\
        \toprule
        Learning rate &  5e-4 \\
        Optimizer & Adam~\citep{adam} \\
        Batch size & 4 \\
        Graph construction criteria & Radius \\
        Graph connection radius & 0.15m \\
        Training sequence length & 25 steps \\
        Training history length & 15 steps \\
        \# graph points per object & 20 \\
        \# graph points per tactile sensor & 20 \\
        Node encoder MLP width & 150 \\
        Node encoder MLP layers & 3 \\
        Edge encoder MLP width & 150 \\
        Edge encoder MLP layers & 3 \\
        Edge effect MLP width & 150 \\
        Edge effect MLP layers & 3 \\
        Edge propagation steps & 3 \\
        Latent physics vector size (dim($\xi$)) & 16 \\
        Tactile encoding dimension (per point in $o^{tact}$) & 5
        
    \end{tabular}
    \caption{\textbf{Hyperparameters for dynamics model training.} We use the same hyperparameters for the nonprehensile box pushing and dense packing tasks.}
    \label{tab:model_training_details}
\end{table}

\section{Hardware Setup}
The hardware setup is depicted in Figure~\ref{fig:hardware} in the main text. 

\textbf{Robot}. We use a Franka Emika Panda robot arm, controlled using the Deoxys open-source controller library~\citep{zhu2022viola}. In our experiments, we use the $\texttt{OSC\_POSITION}$ and $\texttt{OSC\_YAW}$ controllers provided by the Deoxys library.

\textbf{Sensors}. We attach the Soft-Bubble sensors to the Franka Panda gripper using custom-designed 3D-printed adapters. We inflate both Soft-Bubble sensors to a width of 45mm measured from the largest distance sensor frame to the rubber sensor surface. While there can be slight variations in the exact amount of air in the sensor due to measurement error, we do not find this to be a significant cause of domain shift for learned models, likely because the signals that are used as input to our model are largely calculated using differences between the current reading and a reference frame captured when the gripper does not make contact with any object that we reset upon each inflation. While we contribute a novel method for integrating tactile information into the particle-based scene representation, the computation of raw tactile features described in \secref{sec:tactile-perception} are computed by the Soft-Bubble sensor API \citep{kuppuswamy2020softbubble} and is not part of our contribution.

% \section{Planning} \label{appendix:planning}
% In this section, we introduce our model-predictive control method for the box pushing and dense packing tasks.

% %% ST 1/16: Maybe we should move this discussion of action space to the experiments section, as I don't think it's critical to the method section (more of a per-task implementation detail)
% \subsubsection{Action Space}

% \subsubsection{Loss Function} 

\section{Planning Implementation Details}
\label{appendix:planning}
We provide hyperparameters for the MPPI optimizer that is used for planning with learned dynamics models in Table~\ref{tab:planning_details}. We use the same planning hyperparameters for baselines as we do our method.

\begin{table}[htb]
    \centering
    \begin{tabular}{lrr}
        Hyperparameter & Box pushing & Dense packing\\
        \toprule
        History length & 22 & 25 \\
        Action sampler temporal correlation $\beta$* & 0.2 & N/A \\
        MPPI \# action samples & 400 & 150 \\
        MPPI action horizon & 20 & 80 \\
        MPPI \# iterations & 2 & 1 \\
        MPPI scaling temperature $\gamma$* & 100 & N/A \\
        \# steps executed before replanning $K$ & 20 & 45 
         
    \end{tabular}
    \caption{\textbf{Hyperparameters for real world planning experiments.} For the parameters denoted by *, we use the notation from \citet{nagabandi2020deep}. As introduced in Section~\ref{sec:control}, $K$ refers to the number of steps in the best plan found that is actually executed on the real robot before replanning. For box pushing it is the entire plan, while for dense packing it is $45$ out of $80$ steps.}
    \label{tab:planning_details}
\end{table}

\vspace{-15pt}
\section{System Workflow} \label{app:system}

\begin{figure*}[t]
    \centering \includegraphics[width=0.95\textwidth]{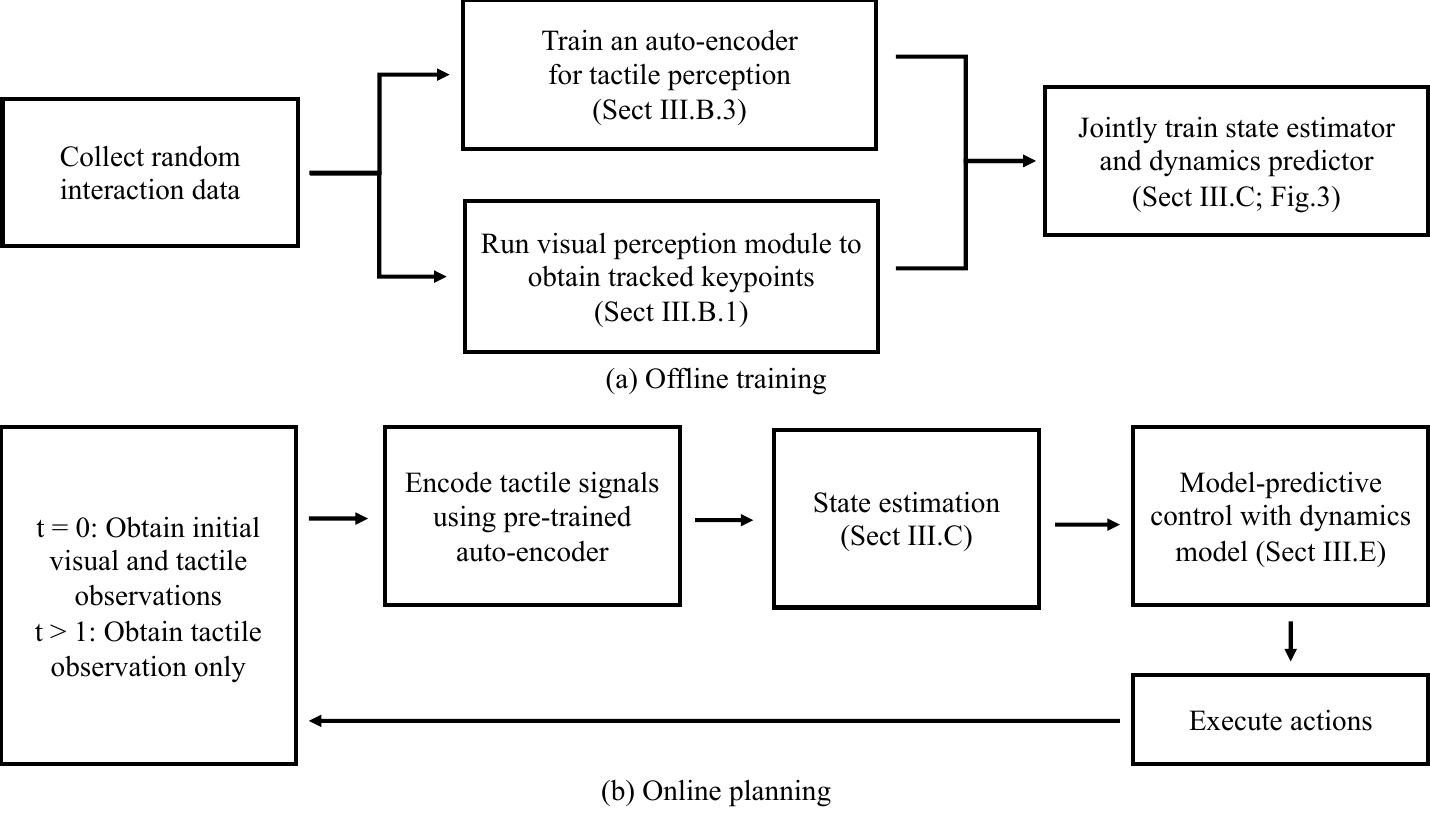}
    \caption{\rebuttal{\textbf{The complete workflow of the RoboPack system.} There are two main stages of the system: (a) offline training and (b) online planning.}}
    \vspace{-0.5cm}
    \label{fig:system}
\end{figure*}

\rebuttal{To present the offline training and online planning processes more clearly, a system diagram is provided in \figref{fig:system}. }

\section{Experiments}

\subsection{Box Configurations}
For the non-prehensile box pushing task, we use boxes that have the same geometry different weight configurations to test the ability of each model to adapt its prediction based on the physical characteristics of each scenario. Specifically, we empty cardboard boxes of dimensions $18\times 9.5 \times 3.8$ cm, and then add metal weights in the following configurations:

% For reference, this is the map of table: our usual box convention:
% Box1 = Box3 
% Box2 = Box7
% Box3 = Box5
% Box4 = Box6 (unseen)

\begin{itemize}
    \item \textbf{Box 1}: Two $100$g weights placed at opposing corners of the box.
    \item \textbf{Box 2}: One $200$g weight placed at the geometric center of the box.
    \item \textbf{Box 3}: No additional weight added.
    \item \textbf{Box 4 (unseen during training)}: This is the original unmodified box, which contains roughly uniformly distributed food items. The items are not affixed to the inner sides of the box, and there could be relative movement between the box and its contents if force is applied.
\end{itemize}

\subsection{Qualitative Results on Planning}
\begin{figure*}[t!]
    \centering
    \includegraphics[width=\linewidth]{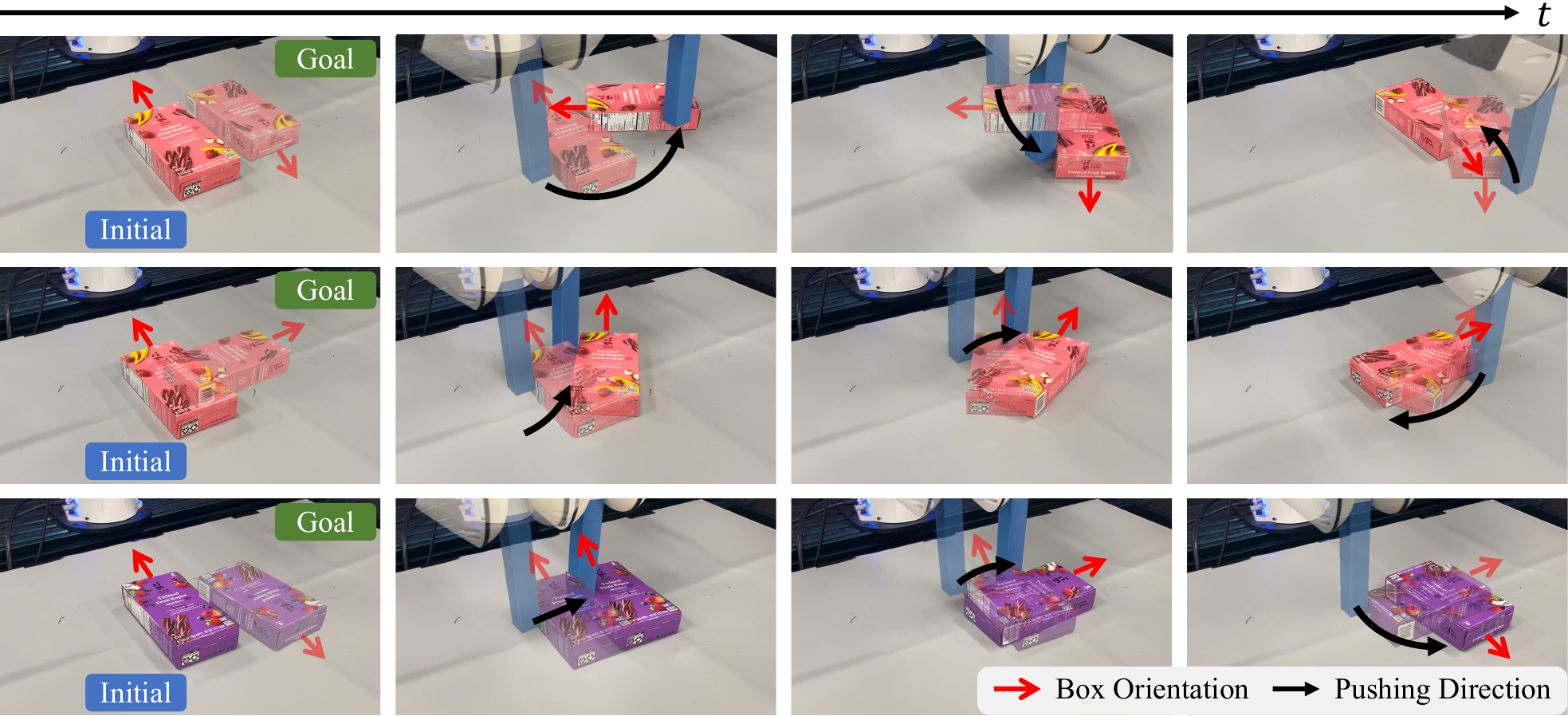} 
    \caption{\textbf{Non-prehensile box pushing.} We demonstrate our robot can push a box with unknown mass distribution from a starting pose to a target pose. Note that our box pushing is non-prehensile because the in-hand object is not fixed. We show that our method can generalize to unseen initial and target box poses in the first two rows and also previously unseen box configurations in the third row. A green arrow indicates the box's orientation, so boxes in rows 1 and 3 are flipped vertically.}
    \label{app:fig:pushing_qualitative}
\end{figure*}

\begin{figure*}[t!]
    \centering
    \includegraphics[width=\linewidth]{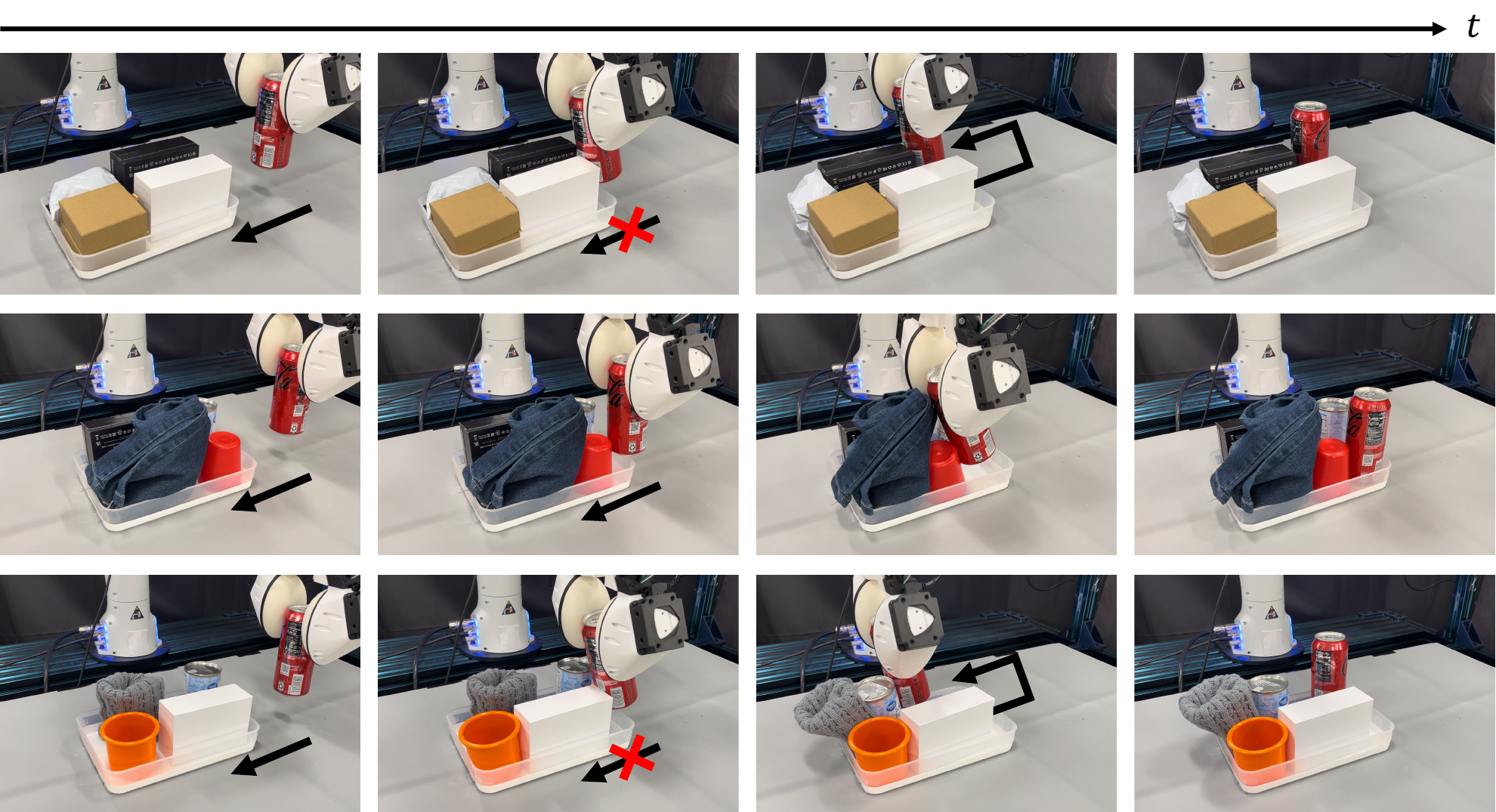} 
    \caption{\textbf{Dense packing with diverse object sets.} In the \textit{Dense Packing} task, we demonstrate that \ours~effectively identifies feasible insertion rows in a tray, minimizing excessive force on the robot to prevent hardware damage. The first row presents a set of objects from data collection, while subsequent rows illustrate our method's capability to adapt to objects with various visual appearances and different levels of deformability.}
    % \vspace{-0.2cm}
    \label{app:fig:packing_qualitative}
\end{figure*}

Additional qualitative results on non-prehensile box pushing and dense packing are presented in 
\figref{app:fig:pushing_qualitative} and \figref{app:fig:packing_qualitative} respectively. Please additionally see our supplementary video for video examples of planning executions.

\subsection{Physics-Based Simulator Baseline}
Here we provide additional details about the physics-based simulator baseline used for the box pushing experiments in Section~\ref{sec:exp}. 

First, we construct a 2D version of the task in the open source Pymunk simulator~\citep{pymunk} that emulates a top-down view of the real scene. The simulated scene contains replicas of the rod and box produced by measuring the dimensions of the real versions of those objects.  

Then, given two visual observations $o_{init}^{vis}$ and $o_{final}^{vis}$(tracked points for each object) and a sequence of actions $\Vec{a}$ taken by the robot, we perform system identification to optimize simulated parameters to fit the real system. Note that our method also uses only two visual observations from the history, but also can use tactile information. Because tactile simulation is not available, the baseline has access to just visual observations. To convert tracked points from real observations into simulator states, we project all points into 2D by truncating the $z$ dimension, and then for each object we compute the object center with the spatial mean of points and the 2D rotation by finding the first two principal components of the 2D points with PCA. Thus the visual observations are converted into tuples $(pos_{init}^{rod}, rot_{init}^{rod}), (pos_{init}^{box}, rot_{init}^{box})$, $(pos_{final}^{rod}, rot_{final}^{rod}),
(pos_{final}^{box}, rot_{final}^{box})$.

We optimize a vector of parameters $\Vec{\mu} \in \mathbb{R}^5$, detailed in Table~\ref{tab:phys_baseline_params}. We de-normalize values from $\Vec{mu}$ to the actual system parameters and clamp them to prevent unrealistic values based on the minimum and maximum values shown. The initial standard deviation for optimization is $\sigma = 0.3$, which we found to work well empirically. The objective function is
$$ \mathcal{L}(\Vec{\mu}) = \|(pos_{final}^{box}, rot_{final}^{box}) - \textsc{sim}_{\Vec{\mu}}(pos_{init}, rot_{init}, \Vec{a}) \|_2. $$

where $\textsc{sim}_{\Vec{\mu}}(pos, rot, \Vec{a})$ represents the box position and rotation after running a simulated trajectory with actions $\Vec{a}$ in the Pymunk simulator starting from box and rod positions $pos$ and rotations $rot$ with simulator parameters set to $\Vec{\mu}$.

We optimize the objective using CMA-ES, a gradient free optimizer, using the implementation from \url{https://github.com/CyberAgentAILab/cmaes}. Parameters are initialized to have the center of mass at the center of the object uniformly distributed mass, and reasonable friction and mass defaults. We use a population size of $8$ based on the implementation-suggested default of $4 + \lfloor 3 * \log (ndim) \rfloor$ and optimize for $100$ generations.  

Finally, we use the optimized set of parameters to perform forward prediction. After forward prediction, we convert the sequence of simulated 2D object positions into a sequence of pointcloud predictions by estimating a rotation matrix and translation (in 2D) and applying them to the 3D pointcloud for the initially provided observation. The z values (height) of all particles are assumed to be fixed at their initial values throughout the prediction. 

\begin{table*}[]
    \centering
    \begin{tabular}{lrrrr}
        Hyperparameter & Initial value & Min& Max & Optimization space $\mu$ to sim. param $p$ transform\\
        \toprule
        Box mass & 10 & 0.001 & N/A & $p = 10(\mu+1)$ \\
        Box friction & 0.5 & 0.0001 & N/A & $p = 0.5(\mu+1)$ \\
        Moment of inertia & 34520.83 & 10 & N/A & $p = 35420.833(\mu+1)$ \\
        Center of gravity x & 0 & -42.5 & 42.5 & $p = 42.5\mu$ \\
        Center of gravity y & 0 & -90 & 90 & $p = 90\mu$ \\ 
    \end{tabular}
    \caption{\textbf{Parameters optimized during system identification for the physics-based simulator baseline.} Initial values and scales are set such that when the parameters in the optimization space are $\Vec{\mu} = 0$, the actual values in the physics simulator $\Vec{p}$ are sensible defaults (see initial value column). Note for center of gravity, $(0, 0)$ refers to the geometric center of the object.}
    \label{tab:phys_baseline_params}
    \vspace{-5pt}
\end{table*}

\section{Tracking Module Details}
As described in Section~\ref{sec:perception}, after sampling initial sets of points for each object $\Vec{p}_{init}$, we formulate point tracking as optimization for the points at each step $\Vec{p}$. Specifically, the new points are computed as a 3D transformation of the points output at the previous step, represented by a rigid rotation $R \in \mathbb{R}^{3}$, translation $T \in \mathbb{R}^3$ and optional per-axis shearing $S \in \mathbb{R}^3$. The transform is a composition of rotation by $R$, scaling by $S$, and translation by $T$ in that order. We abuse notation to sometimes use $\Vec{p}$ for ease of reading, but $\Vec{p}$ is a function of the actual optimized parameters $R, S, T$. Thus the optimization objective has the following loss terms:
\begin{enumerate}
    \item \textbf{Distance to surface.} $$\mathcal{L}_{depth}(\Vec{p}) = \frac{1}{|\Vec{p}|} \sum_{p \in \Vec{p}} \max(0, depth_{interp(p)} - depth_{proj(\Vec{p})}) $$
where $depth_{interp(p)}$ is the depth estimation from interpolating information from multi-view depth observations, and $depth_{proj(\Vec{p})}$ is the expected depth at each point when projected into each camera frame.
    \item \textbf{Semantic alignment.}
     $$\mathcal{L}_{align}(\Vec{p}) = \frac{1}{|\Vec{p}|} \sum_{p \in \Vec{p}} \min(\|dinov2(p_{init}) - dinov2(p)\|_2, 30)$$
     where $dinov2(p)$ represents the multi-view interpolated DinoV2 feature at the 3D point represented by $p$, and again $p_{init}$ is the position of the point in the first frame (not necessarily immediately prior frame) of tracking.
    \item \textbf{Motion regularization.}
    $$ \mathcal{L}_{reg}(R, T, S) = w_{reg}^T \|R\|_2 + w_{reg}^T \|T\|_2 + w_{reg}^S \|S\|_2. $$
    Motion regularization prevents tracked points from exhibiting high frequency jitter when the objects they are tracking do not move. 
    \item \textbf{Mask consistency.}
    We introduce a mask consistency loss. Intuitively, this loss tries to ensure that each pixel within a 2D mask for an object from a particular camera view should have a tracked point for that object that is close to that pixel when projected into that view.

    Let the set of all views be $V$ and the set of object masks in a particular view $v$ be $M(v)$. 
    Then the total number of masks points $N$ is $N = \sum_{v \in V} \sum_{obj \in M(v)} |obj|$.

    Concretely, this can be written as:
    $$\mathcal{L}_{mask}(\Vec{p}) = \frac{1}{N} \sum_{v \in V} \sum_{obj \in M(v)} \sum_{pix \in obj} \min_{p \in \Vec{p}^{obj}} \|pix - proj(p, v)\|$$

    where $proj(p, v)$ is the 2D projection of 3D point $p$ into the image space of viewpoint $v$. 

\end{enumerate}
The overall objective is computed by weighting and combining these terms:
    \[
    \begin{aligned}
        \mathcal{L}_{tracking} &= w_{depth} \mathcal{L}_{depth} 
         +w_{align} \mathcal{L}_{align} +\\ &\quad + w_{reg} \mathcal{L}_{reg} + w_{mask} \mathcal{L}_{mask}
    \end{aligned}
    \]

The weights for each term as well as optimizer parameters are enumerated in Table~\ref{tab:tracking_details}. The transformed points with the best loss after the total number of gradient steps is complete is output as the result.

\begin{table}[]
    \centering
    \begin{tabular}{lrr}
        Hyperparameter & Box pushing & Dense packing\\
        \toprule
        Optimizer & Adam & Adam \\
        LR schedule & Reduce on plateau& Reduce on plateau\\
        Grad steps & 200 & 200 \\
        Learning rate (T) & 0.04 & 0.01 \\
        Learning rate (R) & 0.04 & 0.1 \\
        Learning rate (S) & 0.04 & 0.01 \\
        Use scale term & No & Yes \\
        $w_{depth}$ & 1 & 1 \\
        $w_{align}$ & 1 & 1 \\
        $w_{reg}^T$ & 1e-3 & 3e3 \\
        $w_{reg}^R$ & 1e-3 & 1e2 \\
        $w_{reg}^S$ & N/A & 3e3 \\
        $w_{mask}$ & 100 & 15 \\
    \end{tabular}
    \caption{\textbf{Loss weights for the tracking module.}}
    \label{tab:tracking_details}
    \vspace{-5pt}
\end{table}

\clearpage

%% file: main.bbl
\begin{thebibliography}{63}
\providecommand{\natexlab}[1]{#1}
\providecommand{\url}[1]{\texttt{#1}}
\expandafter\ifx\csname urlstyle\endcsname\relax
  \providecommand{\doi}[1]{doi: #1}\else
  \providecommand{\doi}{doi: \begingroup \urlstyle{rm}\Url}\fi

\bibitem[Ai et~al.(2022)Ai, Gao, Vinay, and Hsu]{ai2022deep}
Bo~Ai, Wei Gao, Vinay, and David Hsu.
\newblock Deep visual navigation under partial observability.
\newblock In \emph{International Conference on Robotics and Automation (ICRA)},
  2022.

\bibitem[Battaglia et~al.(2016)Battaglia, Pascanu, Lai, Jimenez~Rezende, and
  {kavukcuoglu}]{battaglia2016interaction}
Peter Battaglia, Razvan Pascanu, Matthew Lai, Danilo Jimenez~Rezende, and koray
  {kavukcuoglu}.
\newblock Interaction {{Networks}} for {{Learning}} about {{Objects}},
  {{Relations}} and {{Physics}}.
\newblock In \emph{Advances in {{Neural Information Processing Systems}}},
  2016.

\bibitem[Blomqvist(2023)]{pymunk}
Victor Blomqvist.
\newblock {Pymunk}, 2023.
\newblock URL \url{https://pymunk.org}.

\bibitem[Calandra et~al.(2018)Calandra, Owens, Jayaraman, Lin, Yuan, Malik,
  Adelson, and Levine]{calandra2018more}
Roberto Calandra, Andrew Owens, Dinesh Jayaraman, Justin Lin, Wenzhen Yuan,
  Jitendra Malik, Edward~H. Adelson, and Sergey Levine.
\newblock More {{Than}} a {{Feeling}}: {{Learning}} to {{Grasp}} and {{Regrasp
  Using Vision}} and {{Touch}}.
\newblock \emph{IEEE Robotics and Automation Letters}, 2018.

\bibitem[Chang and Pad{\i}r(2020)]{chang2020modelbased}
Peng Chang and Ta{\c s}k{\i}n Pad{\i}r.
\newblock Model-{{Based Manipulation}} of {{Linear Flexible Objects}} with
  {{Visual Curvature Feedback}}.
\newblock In \emph{{{IEEE}}/{{ASME International Conference}} on {{Advanced
  Intelligent Mechatronics}} ({{AIM}})}, 2020.

\bibitem[Chen et~al.(2023)Chen, Niu, Hong, Liu, Wang, Li, and
  {Driggs-Campbell}]{chen2023predictinga}
Haonan Chen, Yilong Niu, Kaiwen Hong, Shuijing Liu, Yixuan Wang, Yunzhu Li, and
  Katherine~Rose {Driggs-Campbell}.
\newblock Predicting {{Object Interactions}} with {{Behavior Primitives}}: {{An
  Application}} in {{Stowing Tasks}}.
\newblock In \emph{{{Conference}} on {{Robot Learning}}}, 2023.

\bibitem[Dahiya et~al.(2010)Dahiya, Metta, Valle, and
  Sandini]{dahiya2010tactile}
Ravinder~S. Dahiya, Giorgio Metta, Maurizio Valle, and Giulio Sandini.
\newblock Tactile {{Sensing}}{\textemdash}{{From Humans}} to {{Humanoids}}.
\newblock \emph{IEEE Transactions on Robotics}, 2010.

\bibitem[Donlon et~al.(2018)Donlon, Dong, Liu, Li, Adelson, and
  Rodriguez]{donlon2018gelslim}
Elliott Donlon, Siyuan Dong, Melody Liu, Jianhua Li, Edward Adelson, and
  Alberto Rodriguez.
\newblock {{GelSlim}}: {{A High-Resolution}}, {{Compact}}, {{Robust}}, and
  {{Calibrated Tactile-sensing Finger}}.
\newblock In \emph{{{IEEE}}/{{RSJ International Conference}} on {{Intelligent
  Robots}} and {{Systems}} ({{IROS}})}, 2018.

\bibitem[Ebert et~al.(2018)Ebert, Finn, Dasari, Xie, Lee, and
  Levine]{visual-foresight}
Frederik Ebert, Chelsea Finn, Sudeep Dasari, Annie Xie, Alex~X. Lee, and Sergey
  Levine.
\newblock Visual foresight: Model-based deep reinforcement learning for
  vision-based robotic control.
\newblock \emph{arXiv:1812.00568}, 2018.

\bibitem[Gao et~al.(2023)Gao, Dou, Li, Agarwal, Bohg, Li, {Fei-Fei}, and
  Wu]{gao2023object}
Ruohan Gao, Yiming Dou, Hao Li, Tanmay Agarwal, Jeannette Bohg, Yunzhu Li,
  Li~{Fei-Fei}, and Jiajun Wu.
\newblock The {{Object Folder Benchmark}} : {{Multisensory Learning}} with
  {{Neural}} and {{Real Objects}}.
\newblock In \emph{{{IEEE}}/{{CVF Conference}} on {{Computer Vision}} and
  {{Pattern Recognition}} ({{CVPR}})}, 2023.

\bibitem[Ha and Schmidhuber(2018)]{world-models}
David Ha and J{\"{u}}rgen Schmidhuber.
\newblock Recurrent world models facilitate policy evolution.
\newblock In Samy Bengio, Hanna~M. Wallach, Hugo Larochelle, Kristen Grauman,
  Nicol{\`{o}} Cesa{-}Bianchi, and Roman Garnett, editors, \emph{Advances in
  Neural Information Processing Systems}, 2018.

\bibitem[Haarnoja et~al.(2018)Haarnoja, Zhou, Abbeel, and Levine]{sac}
Tuomas Haarnoja, Aurick Zhou, Pieter Abbeel, and Sergey Levine.
\newblock Soft actor-critic: Off-policy maximum entropy deep reinforcement
  learning with a stochastic actor.
\newblock In \emph{International Conference on Machine Learning}, 2018.

\bibitem[Hafner et~al.(2019)Hafner, Lillicrap, Fischer, Villegas, Ha, Lee, and
  Davidson]{planet}
Danijar Hafner, Timothy~P. Lillicrap, Ian Fischer, Ruben Villegas, David Ha,
  Honglak Lee, and James Davidson.
\newblock Learning latent dynamics for planning from pixels.
\newblock In Kamalika Chaudhuri and Ruslan Salakhutdinov, editors,
  \emph{International Conference on Machine Learning}, 2019.

\bibitem[Hansen(2016)]{hansen2016cma}
Nikolaus Hansen.
\newblock The cma evolution strategy: A tutorial.
\newblock \emph{arXiv:1604.00772}, 2016.

\bibitem[He(2023)]{offline_model_rl_survey}
Haoyang He.
\newblock A survey on offline model-based reinforcement learning.
\newblock \emph{arXiv:2305.03360}, 2023.

\bibitem[Hester and Stone(2013)]{texplore}
Todd Hester and Peter Stone.
\newblock {TEXPLORE:} real-time sample-efficient reinforcement learning for
  robots.
\newblock \emph{Machine Learning}, 2013.

\bibitem[Holl et~al.(2019)Holl, Thuerey, and Koltun]{holl2019learning}
Philipp Holl, Nils Thuerey, and Vladlen Koltun.
\newblock Learning to {{Control PDEs}} with {{Differentiable Physics}}.
\newblock In \emph{International {{Conference}} on {{Learning
  Representations}}}, 2019.

\bibitem[Kaelbling et~al.(1998)Kaelbling, Littman, and Cassandra]{pomdp_leslie}
Leslie~Pack Kaelbling, Michael~L. Littman, and Anthony~R. Cassandra.
\newblock Planning and acting in partially observable stochastic domains.
\newblock \emph{Artificial Intelligence}, 1998.

\bibitem[Kalashnikov et~al.(2021)Kalashnikov, Varley, Chebotar, Swanson,
  Jonschkowski, Finn, Levine, and Hausman]{mt-opt}
Dmitry Kalashnikov, Jacob Varley, Yevgen Chebotar, Benjamin Swanson, Rico
  Jonschkowski, Chelsea Finn, Sergey Levine, and Karol Hausman.
\newblock Mt-opt: Continuous multi-task robotic reinforcement learning at
  scale.
\newblock \emph{arXiv:2104.08212}, 2021.

\bibitem[Kingma and Ba(2015)]{adam}
Diederik~P. Kingma and Jimmy Ba.
\newblock Adam: {A} method for stochastic optimization.
\newblock In Yoshua Bengio and Yann LeCun, editors, \emph{International
  Conference on Learning Representations}, 2015.

\bibitem[Kirillov et~al.(2023)Kirillov, Mintun, Ravi, Mao, Rolland, Gustafson,
  Xiao, Whitehead, Berg, Lo, Doll{\'a}r, and Girshick]{kirillov2023segany}
Alexander Kirillov, Eric Mintun, Nikhila Ravi, Hanzi Mao, Chloe Rolland, Laura
  Gustafson, Tete Xiao, Spencer Whitehead, Alexander~C. Berg, Wan-Yen Lo, Piotr
  Doll{\'a}r, and Ross Girshick.
\newblock Segment anything.
\newblock \emph{arXiv:2304.02643}, 2023.

\bibitem[Kuppuswamy et~al.(2020)Kuppuswamy, Alspach, Uttamchandani, Creasey,
  Ikeda, and Tedrake]{kuppuswamy2020softbubble}
Naveen Kuppuswamy, Alex Alspach, Avinash Uttamchandani, Sam Creasey, Takuya
  Ikeda, and Russ Tedrake.
\newblock Soft-bubble grippers for robust and perceptive manipulation.
\newblock In \emph{{{IEEE}}/{{RSJ International Conference}} on {{Intelligent
  Robots}} and {{Systems}} ({{IROS}})}, 2020.

\bibitem[Kurniawati et~al.(2008)Kurniawati, Hsu, and Lee]{sarsop}
Hanna Kurniawati, David Hsu, and Wee~Sun Lee.
\newblock {SARSOP:} efficient point-based {POMDP} planning by approximating
  optimally reachable belief spaces.
\newblock In \emph{Robotics: Science and Systems}, 2008.

\bibitem[Kurutach et~al.(2018)Kurutach, Tamar, Yang, Russell, and
  Abbeel]{kurutach2018learning}
Thanard Kurutach, Aviv Tamar, Ge~Yang, Stuart~J Russell, and Pieter Abbeel.
\newblock Learning {{Plannable Representations}} with {{Causal InfoGAN}}.
\newblock In \emph{Advances in {{Neural Information Processing Systems}}},
  2018.

\bibitem[Lambeta et~al.(2020)Lambeta, Chou, Tian, Yang, Maloon, Most, Stroud,
  Santos, Byagowi, Kammerer, Jayaraman, and Calandra]{lambeta2020digit}
Mike Lambeta, Po-Wei Chou, Stephen Tian, Brian Yang, Benjamin Maloon,
  Victoria~Rose Most, Dave Stroud, Raymond Santos, Ahmad Byagowi, Gregg
  Kammerer, Dinesh Jayaraman, and Roberto Calandra.
\newblock {{DIGIT}}: {{A Novel Design}} for a {{Low-Cost Compact
  High-Resolution Tactile Sensor With Application}} to {{In-Hand
  Manipulation}}.
\newblock \emph{IEEE Robotics and Automation Letters}, 2020.

\bibitem[Legaard et~al.(2023)Legaard, Schranz, Schweiger, Drgo{\v n}a, Falay,
  Gomes, Iosifidis, Abkar, and Larsen]{legaard2023constructing}
Christian Legaard, Thomas Schranz, Gerald Schweiger, J{\'a}n Drgo{\v n}a, Basak
  Falay, Cl{\'a}udio Gomes, Alexandros Iosifidis, Mahdi Abkar, and Peter
  Larsen.
\newblock Constructing {{Neural Network Based Models}} for {{Simulating
  Dynamical Systems}}.
\newblock \emph{ACM Computing Surveys}, 2023.

\bibitem[Levine et~al.(2016)Levine, Finn, Darrell, and Abbeel]{e2e_visuomotor}
Sergey Levine, Chelsea Finn, Trevor Darrell, and Pieter Abbeel.
\newblock End-to-end training of deep visuomotor policies.
\newblock \emph{Journal of Machine Learning Research}, 2016.

\bibitem[Li et~al.(2023)Li, Zhang, Zhu, Wang, Lee, Xu, Adelson, {Fei-Fei}, Gao,
  and Wu]{li2023see}
Hao Li, Yizhi Zhang, Junzhe Zhu, Shaoxiong Wang, Michelle~A. Lee, Huazhe Xu,
  Edward Adelson, Li~{Fei-Fei}, Ruohan Gao, and Jiajun Wu.
\newblock See, {{Hear}}, and {{Feel}}: {{Smart Sensory Fusion}} for {{Robotic
  Manipulation}}.
\newblock In \emph{{{Conference}} on {{Robot Learning}}}, 2023.

\bibitem[Li et~al.(2019)Li, Wu, Tedrake, Tenenbaum, and Torralba]{dpi}
Yunzhu Li, Jiajun Wu, Russ Tedrake, Joshua~B. Tenenbaum, and Antonio Torralba.
\newblock Learning particle dynamics for manipulating rigid bodies, deformable
  objects, and fluids.
\newblock In \emph{International Conference on Learning Representations}, 2019.

\bibitem[Li et~al.(2020)Li, Torralba, Anandkumar, Fox, and Garg]{li2020causal}
Yunzhu Li, Antonio Torralba, Animashree Anandkumar, Dieter Fox, and Animesh
  Garg.
\newblock Causal discovery in physical systems from videos.
\newblock In \emph{{{Neural Information Processing Systems}}}, 2020.

\bibitem[Li et~al.(2021)Li, Li, Sitzmann, Agrawal, and Torralba]{li20213d}
Yunzhu Li, Shuang Li, Vincent Sitzmann, Pulkit Agrawal, and Antonio Torralba.
\newblock {{3D Neural Scene Representations}} for {{Visuomotor Control}}.
\newblock In \emph{{{Conference}} on {{Robot Learning}}}, 2021.

\bibitem[Lillicrap et~al.(2016)Lillicrap, Hunt, Pritzel, Heess, Erez, Tassa,
  Silver, and Wierstra]{ddpg}
Timothy~P. Lillicrap, Jonathan~J. Hunt, Alexander Pritzel, Nicolas Heess, Tom
  Erez, Yuval Tassa, David Silver, and Daan Wierstra.
\newblock Continuous control with deep reinforcement learning.
\newblock In \emph{International Conference on Learning Representations}, 2016.

\bibitem[Lin et~al.(2023)Lin, Zhang, Xu, Wu, and Xu]{lin20239dtact}
Changyi Lin, Han Zhang, Jikai Xu, Lei Wu, and Huazhe Xu.
\newblock {9DTact}: A compact vision-based tactile sensor for accurate {3D}
  shape reconstruction and generalizable {6D} force estimation.
\newblock \emph{IEEE Robotics and Automation Letters}, 2023.

\bibitem[Lin et~al.(2022)Lin, Wang, Huang, and Held]{lin2022learning}
Xingyu Lin, Yufei Wang, Zixuan Huang, and David Held.
\newblock Learning {{Visible Connectivity Dynamics}} for {{Cloth Smoothing}}.
\newblock In \emph{{{Conference}} on {{Robot Learning}}}, 2022.

\bibitem[Liu et~al.(2023)Liu, Zeng, Ren, Li, Zhang, Yang, Li, Yang, Su, Zhu,
  et~al.]{liu2023grounding}
Shilong Liu, Zhaoyang Zeng, Tianhe Ren, Feng Li, Hao Zhang, Jie Yang, Chunyuan
  Li, Jianwei Yang, Hang Su, Jun Zhu, et~al.
\newblock Grounding {DINO}: Marrying {DINO} with grounded pre-training for
  open-set object detection.
\newblock \emph{arXiv:2303.05499}, 2023.

\bibitem[Luo et~al.(2018)Luo, Xu, Li, Tian, Darrell, and
  Ma]{luo2018algorithmic}
Yuping Luo, Huazhe Xu, Yuanzhi Li, Yuandong Tian, Trevor Darrell, and Tengyu
  Ma.
\newblock Algorithmic {{Framework}} for {{Model-based Deep Reinforcement
  Learning}} with {{Theoretical Guarantees}}.
\newblock In \emph{International {{Conference}} on {{Learning
  Representations}}}, 2018.

\bibitem[Manuelli et~al.(2020)Manuelli, Li, Florence, and
  Tedrake]{manuelli2020keypoints}
Lucas Manuelli, Yunzhu Li, Pete Florence, and Russ Tedrake.
\newblock Keypoints into the future: Self-supervised correspondence in
  model-based reinforcement learning.
\newblock In \emph{Conference on Robot Learning}, 2020.

\bibitem[Matl and Bajcsy(2021)]{matl2021deformable}
Carolyn Matl and Ruzena Bajcsy.
\newblock Deformable {{Elasto-Plastic Object Shaping}} using an {{Elastic
  Hand}} and {{Model-Based Reinforcement Learning}}.
\newblock In \emph{{{IEEE}}/{{RSJ International Conference}} on {{Intelligent
  Robots}} and {{Systems}} ({{IROS}})}, 2021.

\bibitem[Matsushima et~al.(2021)Matsushima, Furuta, Matsuo, Nachum, and
  Gu]{bremen}
Tatsuya Matsushima, Hiroki Furuta, Yutaka Matsuo, Ofir Nachum, and Shixiang Gu.
\newblock Deployment-efficient reinforcement learning via model-based offline
  optimization.
\newblock In \emph{International Conference on Learning Representations}, 2021.

\bibitem[Mnih et~al.(2015)Mnih, Kavukcuoglu, Silver, Rusu, Veness, Bellemare,
  Graves, Riedmiller, Fidjeland, Ostrovski, Petersen, Beattie, Sadik,
  Antonoglou, King, Kumaran, Wierstra, Legg, and Hassabis]{dqn}
Volodymyr Mnih, Koray Kavukcuoglu, David Silver, Andrei~A. Rusu, Joel Veness,
  Marc~G. Bellemare, Alex Graves, Martin~A. Riedmiller, Andreas Fidjeland,
  Georg Ostrovski, Stig Petersen, Charles Beattie, Amir Sadik, Ioannis
  Antonoglou, Helen King, Dharshan Kumaran, Daan Wierstra, Shane Legg, and
  Demis Hassabis.
\newblock Human-level control through deep reinforcement learning.
\newblock \emph{Nature}, 2015.

\bibitem[Murthy et~al.(2020)Murthy, Macklin, Golemo, Voleti, Petrini, Weiss,
  Considine, {Parent-L{\'e}vesque}, Xie, Erleben, Paull, Shkurti,
  Nowrouzezahrai, and Fidler]{murthy2020gradsim}
J.~Krishna Murthy, Miles Macklin, Florian Golemo, Vikram Voleti, Linda Petrini,
  Martin Weiss, Breandan Considine, J{\'e}r{\^o}me {Parent-L{\'e}vesque}, Kevin
  Xie, Kenny Erleben, Liam Paull, Florian Shkurti, Derek Nowrouzezahrai, and
  Sanja Fidler.
\newblock {{gradSim}}: {{Differentiable}} simulation for system identification
  and visuomotor control.
\newblock In \emph{International {{Conference}} on {{Learning
  Representations}}}, 2020.

\bibitem[Nagabandi et~al.(2020)Nagabandi, Konolige, Levine, and
  Kumar]{nagabandi2020deep}
Anusha Nagabandi, Kurt Konolige, Sergey Levine, and Vikash Kumar.
\newblock Deep {{Dynamics Models}} for {{Learning Dexterous Manipulation}}.
\newblock In \emph{{{Conference}} on {{Robot Learning}}}, 2020.

\bibitem[Oquab et~al.(2023)Oquab, Darcet, Moutakanni, Vo, Szafraniec, Khalidov,
  Fernandez, Haziza, Massa, El{-}Nouby, Assran, Ballas, Galuba, Howes, Huang,
  Li, Misra, Rabbat, Sharma, Synnaeve, Xu, J{\'{e}}gou, Mairal, Labatut,
  Joulin, and Bojanowski]{dinov2}
Maxime Oquab, Timoth{\'{e}}e Darcet, Th{\'{e}}o Moutakanni, Huy Vo, Marc
  Szafraniec, Vasil Khalidov, Pierre Fernandez, Daniel Haziza, Francisco Massa,
  Alaaeldin El{-}Nouby, Mahmoud Assran, Nicolas Ballas, Wojciech Galuba,
  Russell Howes, Po{-}Yao Huang, Shang{-}Wen Li, Ishan Misra, Michael~G.
  Rabbat, Vasu Sharma, Gabriel Synnaeve, Hu~Xu, Herv{\'{e}} J{\'{e}}gou, Julien
  Mairal, Patrick Labatut, Armand Joulin, and Piotr Bojanowski.
\newblock {DINOv2}: Learning robust visual features without supervision.
\newblock \emph{arXiv:2304.07193}, 2023.

\bibitem[Qi et~al.(2023)Qi, Yi, Suresh, Lambeta, Ma, Calandra, and
  Malik]{qi2023general}
Haozhi Qi, Brent Yi, Sudharshan Suresh, Mike Lambeta, Yi~Ma, Roberto Calandra,
  and Jitendra Malik.
\newblock General {{In-hand Object Rotation}} with {{Vision}} and {{Touch}}.
\newblock In \emph{{{Conference}} on {{Robot Learning}}}, 2023.

\bibitem[Rafailov et~al.(2021)Rafailov, Yu, Rajeswaran, and Finn]{lompo}
Rafael Rafailov, Tianhe Yu, Aravind Rajeswaran, and Chelsea Finn.
\newblock Offline reinforcement learning from images with latent space models.
\newblock In Ali Jadbabaie, John Lygeros, George~J. Pappas, Pablo~A. Parrilo,
  Benjamin Recht, Claire~J. Tomlin, and Melanie~N. Zeilinger, editors,
  \emph{Conference on Learning for Dynamics and Control}, 2021.

\bibitem[Rigter et~al.(2022)Rigter, Lacerda, and Hawes]{rambo}
Marc Rigter, Bruno Lacerda, and Nick Hawes.
\newblock {RAMBO-RL:} robust adversarial model-based offline reinforcement
  learning.
\newblock In \emph{Advances in Neural Information Processing Systems}, 2022.

\bibitem[{Sanchez-Gonzalez} et~al.(2020){Sanchez-Gonzalez}, Godwin, Pfaff,
  Ying, Leskovec, and Battaglia]{sanchez-gonzalez2020learning}
Alvaro {Sanchez-Gonzalez}, Jonathan Godwin, Tobias Pfaff, Rex Ying, Jure
  Leskovec, and Peter Battaglia.
\newblock Learning to {{Simulate Complex Physics}} with {{Graph Networks}}.
\newblock In \emph{{{International Conference}} on {{Machine Learning}}}, 2020.

\bibitem[Shi et~al.(2022)Shi, Xu, Huang, Li, and Wu]{shi2022robocraft}
Haochen Shi, Huazhe Xu, Zhiao Huang, Yunzhu Li, and Jiajun Wu.
\newblock {{RoboCraft}}: {{Learning}} to {{See}}, {{Simulate}}, and {{Shape
  Elasto-Plastic Objects}} with {{Graph Networks}}.
\newblock In \emph{Robotics: {{Science}} and {{Systems}}}, 2022.

\bibitem[Shi et~al.(2023{\natexlab{a}})Shi, Xu, Clarke, Li, and
  Wu]{shi2023robocooka}
Haochen Shi, Huazhe Xu, Samuel Clarke, Yunzhu Li, and Jiajun Wu.
\newblock {{RoboCook}}: {{Long-Horizon Elasto-Plastic Object Manipulation}}
  with {{Diverse Tools}}.
\newblock In \emph{{{Conference}} on {{Robot Learning}}}, 2023{\natexlab{a}}.

\bibitem[Shi et~al.(2023{\natexlab{b}})Shi, Xu, Huang, Li, and
  Wu]{shi2023robocraft}
Haochen Shi, Huazhe Xu, Zhiao Huang, Yunzhu Li, and Jiajun Wu.
\newblock {{RoboCraft}}: {{Learning}} to see, simulate, and shape
  elasto-plastic objects in {{3D}} with graph networks.
\newblock \emph{The International Journal of Robotics Research},
  2023{\natexlab{b}}.

\bibitem[Silver and Veness(2010)]{pomcp}
David Silver and Joel Veness.
\newblock Monte-carlo planning in large pomdps.
\newblock In John~D. Lafferty, Christopher K.~I. Williams, John Shawe{-}Taylor,
  Richard~S. Zemel, and Aron Culotta, editors, \emph{Advances in Neural
  Information Processing Systems}, 2010.

\bibitem[Suh et~al.(2022)Suh, Kuppuswamy, Pang, Mitiguy, Alspach, and
  Tedrake]{suh2022seed}
H.J.~Terry Suh, Naveen Kuppuswamy, Tao Pang, Paul Mitiguy, Alex Alspach, and
  Russ Tedrake.
\newblock {{SEED}}: {{Series Elastic End Effectors}} in {{6D}} for
  {{Visuotactile Tool Use}}.
\newblock In \emph{{{IEEE}}/{{RSJ International Conference}} on {{Intelligent
  Robots}} and {{Systems}} ({{IROS}})}, 2022.

\bibitem[Suresh et~al.(2023)Suresh, Qi, Wu, Fan, Pineda, Lambeta, Malik,
  Kalakrishnan, Calandra, Kaess, Ortiz, and Mukadam]{suresh2023neural}
Sudharshan Suresh, Haozhi Qi, Tingfan Wu, Taosha Fan, Luis Pineda, Mike
  Lambeta, Jitendra Malik, Mrinal Kalakrishnan, Roberto Calandra, Michael
  Kaess, Joseph Ortiz, and Mustafa Mukadam.
\newblock {N}eural feels with neural fields: {V}isuo-tactile perception for
  in-hand manipulation.
\newblock \emph{arXiv:2312.1346}, 2023.

\bibitem[Tian et~al.(2019)Tian, Ebert, Jayaraman, Mudigonda, Finn, Calandra,
  and Levine]{tian2019manipulation}
Stephen Tian, Frederik Ebert, Dinesh Jayaraman, Mayur Mudigonda, Chelsea Finn,
  Roberto Calandra, and Sergey Levine.
\newblock Manipulation by feel: Touch-based control with deep predictive
  models.
\newblock In \emph{International Conference on Robotics and Automation (ICRA)},
  2019.

\bibitem[Wang et~al.(2023{\natexlab{a}})Wang, Li, {Driggs-Campbell}, {Fei-Fei},
  and Wu]{wang2023dynamicresolutiond}
Yixuan Wang, Yunzhu Li, Katherine {Driggs-Campbell}, Li~{Fei-Fei}, and Jiajun
  Wu.
\newblock Dynamic-{{Resolution Model Learning}} for {{Object Pile
  Manipulation}}.
\newblock In \emph{Robotics: {{Science}} and {{Systems}}}, 2023{\natexlab{a}}.

\bibitem[Wang et~al.(2023{\natexlab{b}})Wang, Li, Zhang, Driggs-Campbell, Wu,
  Fei-Fei, and Li]{wang2023d3fields}
Yixuan Wang, Zhuoran Li, Mingtong Zhang, Katherine Driggs-Campbell, Jiajun Wu,
  Li~Fei-Fei, and Yunzhu Li.
\newblock D$^3$fields: Dynamic 3d descriptor fields for zero-shot generalizable
  robotic manipulation.
\newblock \emph{arXiv:2309.16118}, 2023{\natexlab{b}}.

\bibitem[Weinstein et~al.(2006)Weinstein, Teran, and
  Fedkiw]{weinstein2006dynamic}
R.~Weinstein, J.~Teran, and R.~Fedkiw.
\newblock Dynamic simulation of articulated rigid bodies with contact and
  collision.
\newblock \emph{IEEE Transactions on Visualization and Computer Graphics},
  2006.

\bibitem[Williams et~al.(2016)Williams, Drews, Goldfain, Rehg, and
  Theodorou]{mppi2016}
Grady Williams, Paul Drews, Brian Goldfain, James~M. Rehg, and Evangelos~A.
  Theodorou.
\newblock Aggressive driving with model predictive path integral control.
\newblock In \emph{International Conference on Robotics and Automation (ICRA)},
  2016.

\bibitem[Wu et~al.(2022)Wu, Escontrela, Hafner, Abbeel, and
  Goldberg]{daydreamer}
Philipp Wu, Alejandro Escontrela, Danijar Hafner, Pieter Abbeel, and Ken
  Goldberg.
\newblock Daydreamer: World models for physical robot learning.
\newblock In \emph{Conference on Robot Learning}, 2022.

\bibitem[Yuan et~al.(2017)Yuan, Dong, and Adelson]{yuan2017gelsighta}
Wenzhen Yuan, Siyuan Dong, and Edward~H. Adelson.
\newblock {{GelSight}}: {{High-Resolution Robot Tactile Sensors}} for
  {{Estimating Geometry}} and {{Force}}.
\newblock \emph{Sensors}, 2017.

\bibitem[Yuan et~al.(2023)Yuan, Che, Qin, Huang, Yin, Lee, Wu, Lim, and
  Wang]{yuan2023robot}
Ying Yuan, Haichuan Che, Yuzhe Qin, Binghao Huang, Zhao-Heng Yin, Kang-Won Lee,
  Yi~Wu, Soo-Chul Lim, and Xiaolong Wang.
\newblock Robot {{Synesthesia}}: {{In-Hand Manipulation}} with {{Visuotactile
  Sensing}}.
\newblock \emph{arXiv:2312.01853}, 2023.

\bibitem[Zhang et~al.(2019)Zhang, Vikram, Smith, Abbeel, Johnson, and
  Levine]{solar}
Marvin Zhang, Sharad Vikram, Laura~M. Smith, Pieter Abbeel, Matthew~J. Johnson,
  and Sergey Levine.
\newblock {SOLAR:} deep structured representations for model-based
  reinforcement learning.
\newblock In \emph{International Conference on Machine Learning}, 2019.

\bibitem[Zhu et~al.(2022)Zhu, Joshi, Stone, and Zhu]{zhu2022viola}
Yifeng Zhu, Abhishek Joshi, Peter Stone, and Yuke Zhu.
\newblock {VIOLA}: Imitation learning for vision-based manipulation with object
  proposal priors.
\newblock In \emph{Conference on Robot Learning}, 2022.

\end{thebibliography}
